\documentclass[12pt]{article}
\usepackage{comment}
\usepackage{makecell}
\usepackage{float}
\usepackage{placeins}
\usepackage{algorithm}
\usepackage{algcompatible}
\usepackage{algpseudocode}
\usepackage[table]{xcolor}  
\usepackage{colortbl}       
\definecolor{lightgray}{gray}{0.9}
\usepackage{multicol}
\usepackage{bm}
\usepackage{gensymb}
\usepackage{siunitx}
\usepackage{hyperref}
\usepackage[caption=false,font=normalsize,labelfont=sf,textfont=sf]{subfig}
\usepackage[labelsep=period]{caption}  

\newenvironment{sciabstract}{%
\begin{quote} \bf}
{\end{quote}}

\newcounter{movie}
\renewcommand{\themovie}{S\arabic{movie}}
\newcommand{\movie}[3][]{%
  \refstepcounter{movie}%
  \begin{center}
    \textbf{Movie \themovie. #2} #3
  \end{center}
  \label{movie:#1}%
}

\usepackage{newtxtext,newtxmath}

\usepackage{graphicx}

\usepackage[letterpaper,margin=1in]{geometry}

\renewenvironment{abstract}
	{\quotation}
	{\endquotation}

\date{}

\makeatletter
\renewcommand{\fnum@figure}{\textbf{Fig. \thefigure}}
\renewcommand{\fnum@table}{\textbf{Table \thetable}}
\makeatother

\usepackage{scicite}

\usepackage{url}

\def\scititle{
Curiosity-Driven Development of Action and Language in Robots Through Self-Exploration

Short title: 

Curiosity-Driven Development of Action and Language
}
\title{\bfseries \boldmath \scititle}

\author{
	Theodore J. Tinker$^{1}$,
        Kenji Doya$^{1}$,
	Jun Tani$^{1*}$\and
	\small$^{1}$ Okinawa Institute of Science and Technology, Okinawa, Japan.\and
     \text{*} To whom correspondence should be addressed; E-mail: jun.tani@oist.jp.
}

\begin{document} 

\maketitle

Teaser: Robots learn to perform actions associated with sentences via curiosity-driven self-exploration


\begin{abstract} \bfseries \boldmath

\center{Abstract}

\begin{sciabstract}
Infants acquire language with generalization from minimal experience, whereas large language models require billions of training tokens. What underlies efficient development in humans? We investigated this problem through experiments wherein robotic agents learn to perform actions associated with imperative sentences (e.g., push red cube) via curiosity-driven self-exploration. Our approach amortizes active inference using Q-learning, enabling intrinsically motivated developmental learning. The simulations reveal key findings corresponding to observations in developmental psychology.
i) Generalization improves drastically as the scale of compositional elements increases.
ii) Curiosity-driven exploration enables faster learning.
iii) Rote pairing of sentences and actions precedes compositional generalization.
iv) Exception-handling induces U-shaped developmental performance, a pattern like representational redescription in child language learning.
These results suggest that curiosity-driven active inference accounts for how intrinsically motivated sensorimotor–linguistic learning supports scalable compositional generalization and exception handling in humans and artificial agents.
\end{sciabstract}

\end{abstract}

\section*{Introduction}

\noindent
A central question in both cognitive science and artificial intelligence is how humans and artificial systems can acquire competencies for language and action developmentally, despite having access to only limited learning experiences. This question is exemplified in human infants, who achieve remarkable generalization with sparse input. This is a stark contrast to large-scale models which rely on massive training corpora to reach similar capabilities. This raises the issue of what mechanisms enable such efficient developmental learning.

\textbf{Compositionality, generalization, and exception handling:}
From the perspective of developmental psychology, infants acquire language through rich interaction with their embodied environments. Tomasello's ``verb-island'' hypothesis argues that children initially learn verbs in specific, isolated contexts before generalizing across broader linguistic structures with compositionality \cite{tomasello2003constructing}. 
He also emphasized the importance of embodiment in language acquisition, suggesting that grounding linguistic symbols in sensorimotor experiences is fundamental to language learning \cite{tomasello2003constructing}. 
This view aligns with other studies in developmental psychology highlighting the role of compositionality and generalization in language acquisition \cite{gleitman1990structural, bloom2000how, smith2005development}.

In linguistic theory, \emph{compositionality} refers to the principle that the meaning of a complex expression is determined by the meanings of its constituent parts together with the rules used to combine them \cite{Frege1892English,Montague1970,Szabo2017}. 
This principle implies that linguistic representations are structured and rule-governed, typically exhibiting hierarchical organization. 
A key cognitive consequence of compositional structure is \emph{systematicity}, namely the capacity to understand and produce novel but structurally related expressions by recombining familiar elements \cite{FodorPylyshyn1988}. 
In this view, systematic generalization across combinations of verbs, adjectives, and nouns reflects the presence of an underlying compositional organization, rather than mere associative pairing. 
Thus, generalization is not itself the definition of compositionality, but a behavioral manifestation of compositional representations, enabling learners to flexibly construct and interpret utterances that have not been directly encountered \cite{LakeBaroni2018}.

In the present study, compositionality is operationalized in terms of such systematic generalization across novel combinations of known elements \cite{FodorPylyshyn1988,LakeBaroni2018}. 
We emphasize that this operationalization captures only one aspect of compositionality, and does not aim to model the full hierarchical and formal semantic structure of natural language as defined in classical linguistic theory \cite{Frege1892English,Montague1970}. 
Rather, our focus is on the developmental emergence of systematic recombination capacity in embodied agents as a minimal and tractable step toward understanding the origins of compositionality.
A nontrivial problem is that although the number of possible compositions grows multiplicatively with the vocabulary size (i.e., number of verbs $\times$ number of adjectives $\times$ number of nouns), children achieve generalization after experiencing only a small subset of learning examples. 
This suggests that the effective sample complexity could be proportional to the sum of elements rather than their product. This phenomenon is closely related to the ``poverty of the stimulus'' problem articulated by Chomsky \cite{chomsky1980rules}, which asks how learners generalize so effectively given severely sparse input.

Beyond these, it is well known that children can develop the capacity for \textit{exception-handling}, a hallmark of flexible cognition. In human development, exceptions such as irregular verbs or inconsistent mappings often produce non-monotonic, U-shaped learning trajectories: children first apply a correct form, then overgeneralize it (producing errors), and finally recover the correct rule. This pattern has been widely interpreted as evidence of internal representational reorganization or \textit{representational redescription} \cite{karmiloffsmith1992beyond}. 
Computationally, such U-shaped performance has been demonstrated in models of language acquisition and rule learning \cite{rumelhart1986pasttense, plunkett1991ushape, marchman1993acquisition, elman1996rethinking, mareschal2001computational}. Developmentally, these phenomena reflect the tension between rote memorization, generalization, and the later refinement of exception rules.

\textbf{Synthetic robotics study using active inference:} 
How can humans develop capacity for compositionality as systematic generalization even with exception-handling through learning from sparse input?
To investigate this question, one promising approach is to reconstruct developmental learning processes in machines and robots. The field of developmental robotics has long pursued this line of research, aiming to replicate human-like learning trajectories in embodied systems \cite{asada2001cognitive, kuniyoshi2006early, sandini2007developmental, prescott2024synthesizing}. However, relatively few studies have focused on development of language and motor control under conditions of stimulus poverty. Existing work has primarily examined associative mappings between linguistic input and motor commands in one-shot or supervised batch learning schemes \cite{cangelosi2004simulation, sugita2005cross, taniguchi2016spatial, vijayaraghavan2021grounding}. These approaches neglect the self-directed, developmental context of infant learning.


In this study, we propose a novel scheme for self-exploratory learning of robots by integrating active inference \cite{friston2011, pezzulo2018hierarchical, parr_friston_2019} and reinforcement learning \cite{WatkinsDayan1992,SuttonBarto2018}.
Our integration of active inference and reinforcement learning can be regarded as
amortizing inference, sometimes known as learning to infer or deep active inference
\cite{fountas2020deep, millidge2020deep, ueltzhoffer2018deep}.
However, we have gone further than usual amortization schemes by replacing extrinsic reward alone with expected free energy, which includes expected information gain. This means agents are effectively rewarded for being curious, or, more simply, for learning to be curious.
Our approach to integrate reinforcement learning with active inference was originally inspired by the work of Kawahara et al. \cite{kawahara}. In our model, originally introduced in \cite{tinker2024active}, motor commands are reinforced by two intrinsic rewards: curiosity (seeking unpredictable sensory consequences) and motor entropy (seeking random movements). Motor commands are also reinforced by extrinsic rewards for successfully achieving goals specified by given imperative sentences. Importantly, our previous experiments in maze navigation demonstrated that the combination of curiosity and motor entropy is crucial for enhancing self-exploration, as agents achieved significantly improved exploratory behaviors under this dual-intrinsic reward scheme. Our approach aligns with broader research on self-exploration in machine learning, in which agents are intrinsically rewarded for taking motor actions whose consequences are unpredictable and therefore lead to the greatest information gain, i.e., have epistemic value \cite{oudeyer2007intrinsic, schmidhuber1991possibility}. 

\begin{figure}[ht!]
    \centering
    \includegraphics[width=1\textwidth]{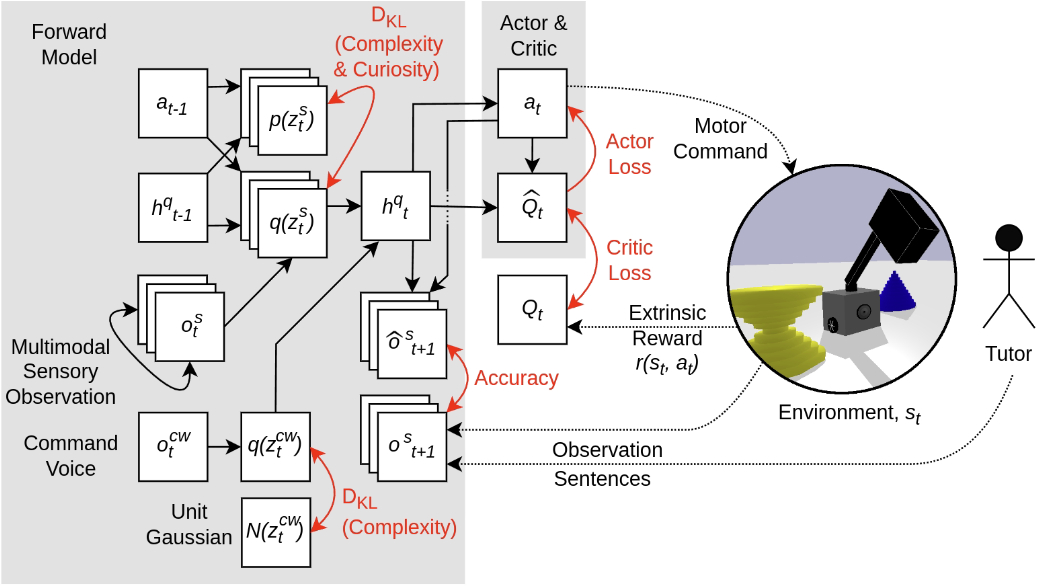}
    \caption{\textbf{The proposed model architecture.}
    The model consists of a forward model, actor, and critic. Variables include: world state $s_t$, sensory observation $o^s_t$, command voice $o^{cw}_t$, motor command $a_t$, Q-value $Q_t$, latent control variables $h^q_t$, random latent variables $z_t$, prior for sensations $p(z^s_t)$, approximated posterior for sensations $q(z^s_t)$, and approximated posterior for command voice $q(z^{cw}_t)$.}
    
    \label{fig:models}
\end{figure}

A simulated mobile robot equipped with a manipulator arm, vision sensor, and distributed tactile sensors learns to generate motor movements in response to imperative sentences (command voice) presented during each trial. These sentences are systematically composed of verbs, adjectives, and nouns, enabling evaluation of generalization performance under different levels of compositional complexity. 
The model architecture employed in this study is based on our previous work \cite{tinker2024intrinsic} with key modifications to accommodate multi-modal sensorimotor integration. (See details in the Materials and Methods section.)
Fig. \ref{fig:models} presents the model architecture, which is composed of three main components: a forward model, an actor part, and a critic part.
The forward model learns to predict the next sensation $o^s_{t+1}$ based on the current sensation $o^s_{t}$, command voice $o^{cw}_{t}$, and the executed motor command $a_{t}$. 
The sensation includes pixel-based vision, tactile sensation, and arm joint proprioception.
To address the hidden state problem and probabilistic nature of the environment, the prediction is performed contextually and stochastically using the random latent variables $z^s_t$ and latent control variables $h^q_t$ wherein the former with gaussian probability distribution is inferred by minimizing the variational free energy while the latter is just computed forwardly using sampling of the inferred random latent variable.
Here, $p(z^s_t)$ is the prior probabilistic distribution of the random latent variables before observing the current sensation while $q(z^s_t)$ is the posterior one after observation of the current sensation caused by the current motor command execution, as will be detailed later.
The latent control variables were shared for each sensory modality, while the random latent variables were allocated separately.
This separation of random latent variables was necessary for the system to deal with different types of sensory modalities simultaneously.
The actor module generates the next motor command $a_{t}$ based on the latent control variable $h^q_t$, which integrates current sensation $o^s_{t}$. 
Based on $a_{t}$ and $h^q_t$, the critic generates $\widehat{Q}$, which is a prediction of the $Q$ value defined with Eq. \ref{eq:bellman_plus_curiosity_method}. While the critic learns to make accurate predictions about future rewards in $Q$, the actor learns to produce motor commands which maximize the critic's predictions $\widehat{Q}$. 
(In the actual architecture there is another sensory input, so-called tutor-feedback voice. This additional input will be detailed later.)

The overall flow is: with a command sentence given by the tutor, the robot attempts to achieve the specified goal by generating a sequence of motor commands. Meanwhile, the forward model predicts the next sensation by inferring the posterior probability distribution $q(z^s_{t} |o_{t},h_{t-1})$ of the random latent variable $z^s$.
This inference is conducted by minimizing the variational free energy $F$ (Eq. \ref{eq:F_intro}). This consists of the complexity term represented by Kullback–Leibler divergence (KLD) between the approximated posterior and the prior, and the accuracy term under the free energy principle (FEP). (See more details on the variational free energy and expected free energy in the ``Free Energy Principle, Active Inference, and Kawahara Model'' section of the Supplementary Materials.)

\begin{align} \label{eq:F_intro}
F_{\psi,t} = \underbrace{D_{KL}[q(z_{t} |o_{t},h_{t-1})||p(z_t|h_{t-1})]}_{\text{Complexity}} - \underbrace{\mathbb{E}_{q(z_t)}[\log p(o_{t+1}|h_t)]}_{\text{Accuracy}}.
\end{align}

\noindent
The forward model is trained iteratively by optimizing its learning parameters $\psi$ in the direction of minimizing the evidence free energy.
The actor learns to generate motor command sequences in the direction of minimizing the expected free energy $G$ (Eq. \ref{eq:G_intro}) through reinforcement learning. 
This consists of the complexity term, extrinsic reward term, and the motor entropy term.
\begin{align} \label{eq:G_intro}
G(a_t) &=  -\underbrace{D_{KL}[q(z_{t}|o_{t},h_{t-1})||p(z_{t}|h_{t-1})]}_{\text{Curiosity}} - \underbrace{r(s_t, a_t)}_{\text{Extrinsic Reward}} - \underbrace{\mathcal{H}(\pi_\phi(a_t|h_{t}))}_{\text{Entropy}} 
\end{align}

It is interesting to note that minimizing evidence free energy $F$ minimizes the complexity term, while minimizing expected free energy $G$ maximizes the same complexity term.
This means that motor commands are generated in the direction of maximizing the information gain attained after execution of the motor command, which is represented by KLD between the approximated posterior and the prior. 
This generates curiosity-driven exploration wherein the agent seeks out previously un-encountered sensorimotor experiences.
On the other hand, the learning parameters in the forward model are updated in the direction of minimizing the same complexity term, representing the latent conflict that is generated by novel experiences encountered during curiosity-driven exploration.
Therefore, both processes of self-exploration and the forward model learning are complementing each other as if in competition. In other words, inference about latent states causing observations is trying to minimize complexity, while learning which actions to take is trying to maximize complexity expected after acting.

Note also that the expected information gain is evaluated using observations at each time point. This can be compared with the usual expressions for expected free energy in which an expectation over future observations is used. In general, observations in the future are random variables and the sign of the KLD subtending complexity switches so that expected information gain is maximized. However, because we are amortizing or learning to minimize expected free energy, we can evaluate the expected complexity based upon current observations, thereby enabling the agent to learn to be information seeking. In other words, expected free energy is not used to select the next action, it is used as an objective function to learn what action to take next.

\textbf{Limitations:} Although the present study draws inspiration from developmental psychology, the resemblance between our model and human infant language development remains limited. Human infants undergo extensive sensorimotor and social learning for at least the first six months of life before exhibiting significant linguistic competence \cite{Kuhl2004,AdolphFranchak2015}. Early language acquisition typically progresses from single-word utterances to two-word combinations and only later to more complex multi-word constructions \cite{bloom2000how,tomasello2003constructing}. Around the onset of combinatorial speech, infants often exhibit a vocabulary spurt, acquiring new words at a remarkably rapid pace \cite{GoldfieldReznick1990}. These developmental milestones occur within a richly scaffolded social environment, including joint attention, imitation, and caregiver feedback, which substantially reduce the complexity of the learning problem \cite{Carpenter1998,Bruner1983,vygotsky1978mind}. In contrast, our model operates in a simplified environment without prior non-linguistic developmental stages, social scaffolding, or progressive linguistic complexity. 

Furthermore, the present framework assumes learning from scratch without incorporating innate cognitive biases, architectural constraints, or evolutionary priors that are widely argued to support human language and action acquisition \cite{chomsky1980rules,Pinker1994,Spelke2007}. Human learners likely benefit from species-specific perceptual, social, and representational predispositions shaped by evolutionary history. Our model does not attempt to capture such innate structures. Instead, it investigates how compositionality as systematic generalization can emerge from curiosity-driven sensorimotor learning under minimal assumptions. Therefore, while the present results demonstrate a mechanistic account of systematic recombination in embodied agents, they should not be interpreted as a comprehensive model of human language development. 
Rather, they provide a simplified testbed for examining how associative learning of language and action based on active inference may contribute to the emergence of structured generalization.

\textbf{Hypothesis:} 
This study therefore tested the following hypotheses through simulation experiments:  
\textbf{H1:} Generalization performance improves drastically as the scale of compositionality in the task increases.  
\textbf{H2:} Curiosity augmented with motor entropy enhances the performance of developmental learning.  
\textbf{H3:} In the early phase, actions are generated only for exactly learned imperative sentences, but in later phases, the system generalizes to novel, unlearned compositions.  
\textbf{H4:} Primitive actions are acquired earlier, followed by more complex, prerequisite-dependent actions.  
\textbf{H5:} Exception-handling rules can be acquired through exploratory learning, exhibiting U-shaped developmental performance similar to that observed in human development.

\section*{Results}
\subsection*{Task Description}
\begin{figure}[ht!]
    \centering
    \includegraphics[width=.75\textwidth]{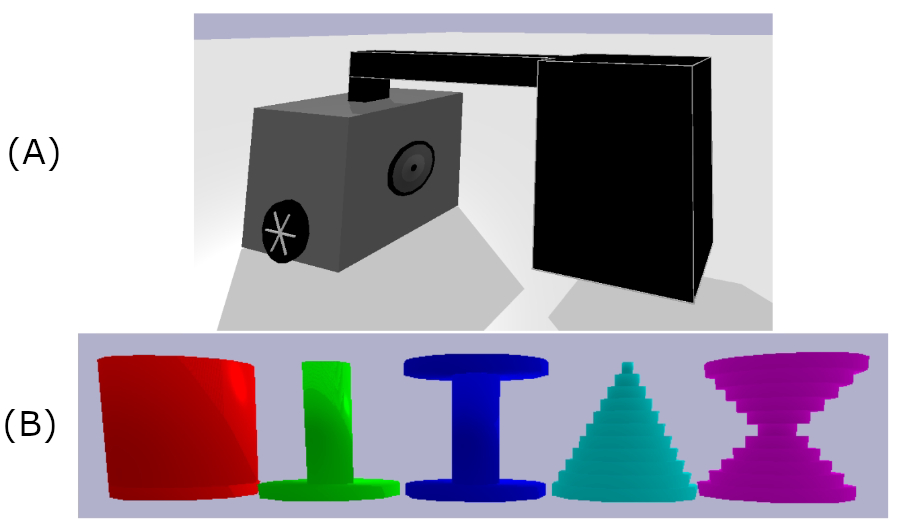}
    \caption{\textbf{The simulated robot and a set of objects to act on}. \textbf{(A)} The robot has two wheels and an arm with two joints. The design is similar to a truck crane. \textbf{(B)} Left to right: a red pillar, a green pole, a blue dumbbell, a cyan cone, and a magenta hourglass. The color yellow is not pictured here.}
    \label{fig:robot}
\end{figure}

We created a robot like a truck crane in a physics simulator along with a set of objects with 5 different shapes each of which can take 6 different colors (see Fig. \ref{fig:robot}).
The robot can maneuver by controlling velocity of left and right wheels independently, and also can move its arm by controlling rotation velocity of the yaw and pitch joint angles for acting on the objects. A camera with 16 x 16 pixels was fixed to the body for visual sensation. 16 touch sensors were distributed in the body and the arm, and rotation angles for the yaw and pitch were sensed as proprioception.

For each trial episode, a task goal was given in terms of an imperative sentence composed with verb, adjective, and noun. Possible words used for them are shown in Table 1. 
\begin{table}[h]
\centering
\begin{tabular}{|c|c|c|}
\hline
\multicolumn{3}{|c|}{\textbf{English Words}} \\
\hline
\multicolumn{1}{|c|}{\textbf{Verb}} & 
\multicolumn{1}{|c|}{\textbf{Adjective}} & 
\multicolumn{1}{|c|}{\textbf{Noun}} \\
\hline
watch &
red & 
pillar \\
be near & 
green &
pole \\

touch the top & 
blue &
dumbbell \\

push forward & 
cyan & 
cone \\

push left & 
magenta & 
hourglass \\

push right & 
yellow & 
    \\

\hline
\end{tabular}
\caption{\textbf{English words.} The English words used for imperative sentences specifying goals.}
\label{table:english_words}
\end{table}
In the beginning of each episode, two objects were located at random positions in the arena wherein one object was the one specified in the imperative sentence and the other was the one with randomly selected color and shape combination among possible ones. 

At each step, the robot receives visual sensation, proprioception for the arm, tactile sensation, 
and two types of voices: the command voice and the tutor-feedback voice. 
The command voice takes the format of the imperative sentence described previously, and is delivered continuously at every step from the beginning.
On the other hand, the feedback voice arrives whenever the robot achieves one of possible goals even if the achieved goal is not the imperative sentence told by the command voice, and it informs which goal has been actually achieved in the same format with the command voice.
This potentially enhances the forward model's learning about its own action as associated with linguistic representation. 
Finally, when the goal specified by the command voice is achieved, a reward is provided.
Each trial episode ran for 30 steps, or was terminated when the specified goal is achieved.
Then the robot is trained with a random batch of 32 episodes.

\subsection*{Effects of curiosity: Experiment 1}
This experiment examined effects of different levels of curiosity to the developmental learning processes using the basic setup.
In the basic setup, full compositions of words (Table \ref{table:english_words}) were used to generate the imperative sentences.
However, the training was conducted using only 60 imperative sentences (33 percent) out of 180 possible sentences.
120 untrained sentences were used for generalization testing.
For ten robots with different random seeds, the complete developmental learning process was iterated for 60,000 epochs. 
The generalization testing with unlearned imperative sentences was conducted every 50 epochs.

The experiment was conducted by changing the levels of curiosity.
Since the random latent variables are computed separately for each sensory modality, the complexity or curiosity can be computed for each sensory modality.
Three levels of curiosity were considered in computing expected free energy $G$: \textit{no curiosity}, wherein none of the curiosity terms for sensory modalities are included; \textit{sensory-motor curiosity}, wherein the curiosity terms only for vision, tactile sensation, and proprioception are included; and \textit{all curiosity}, wherein the curiosity terms for all sensory modalities including feedback voice are included.

\begin{figure}[ht!]
    \centering
    \includegraphics[width=\textwidth]{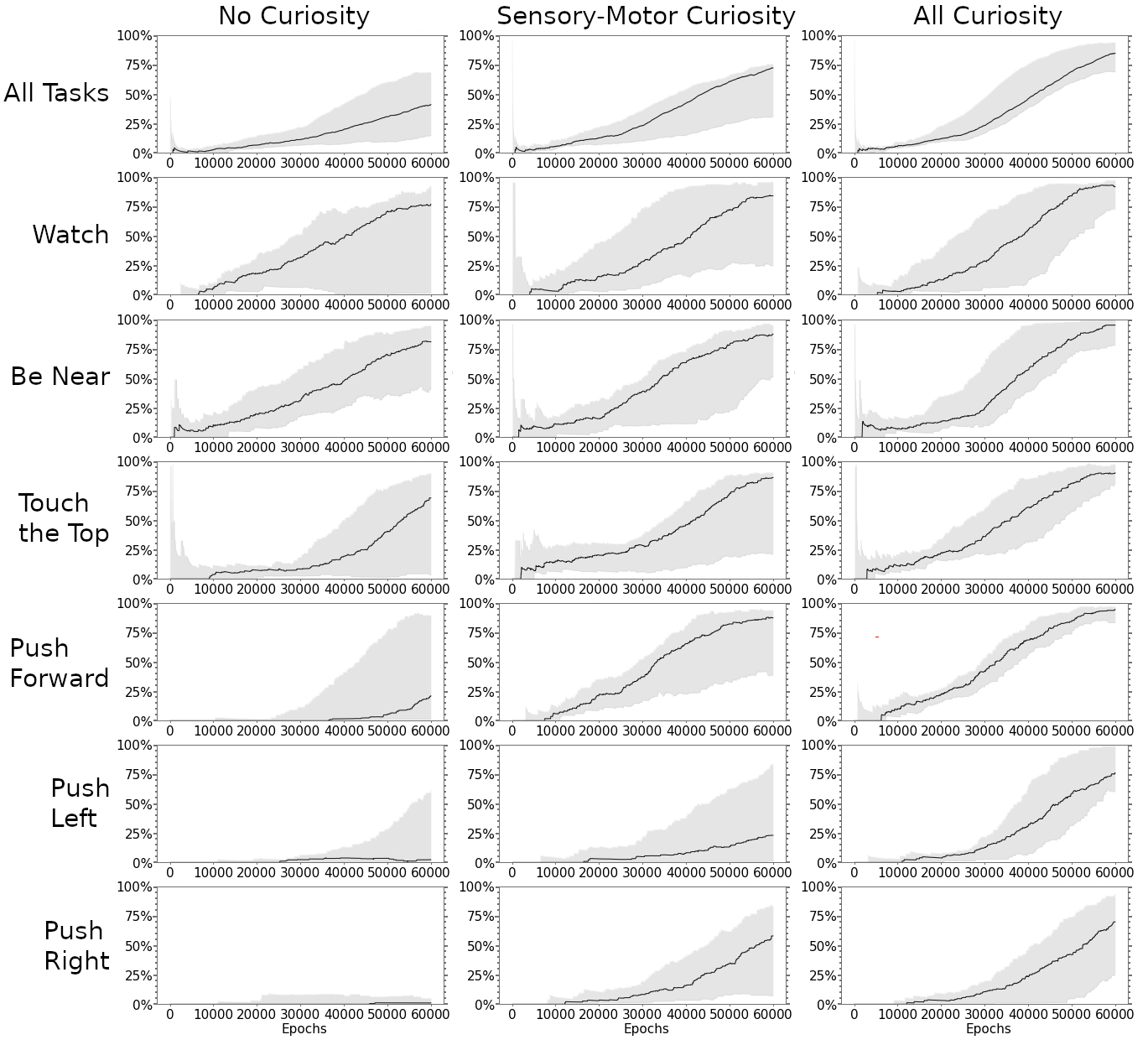}
    \caption{\textbf{Rolling success-rates for unlearned goals.} Compares agents with different levels of curiosity. Shaded areas represent 99\% confidence intervals for the 10 agents.}
    \label{fig:evaluation}
\end{figure}

Fig. \ref{fig:evaluation} shows the development of the generalization testing performances in terms of success rate for goals specified by unlearned imperative sentences, which are plotted for different action categories with different levels of curiosity. 
The plots show that the performance was improved significantly as the curiosity level was increased.
Importantly, the case of all actions with the \textit{all curiosity} level shows that the average success rate for unlearned goals reached a quite high value of 85.1 percent even though the learning was conducted only for 33 percent of all possible compositions.

It can also be observed in Fig. \ref{fig:evaluation} that some action categories develop more slowly than others in the \textit{all curiosity} condition. In particular, ``push right'' and ``push left'' exhibit slower development compared to the other action categories. This may be attributed to the relatively higher behavioral complexity associated with these actions. In contrast, no clear or significant differences in development speed are observed among the remaining actions, namely ``watch,'' ``be near,'' ``push forward,'' and ``touch the top.''

\begin{figure}[ht!]
    \centering
    \includegraphics[width=.75\textwidth]{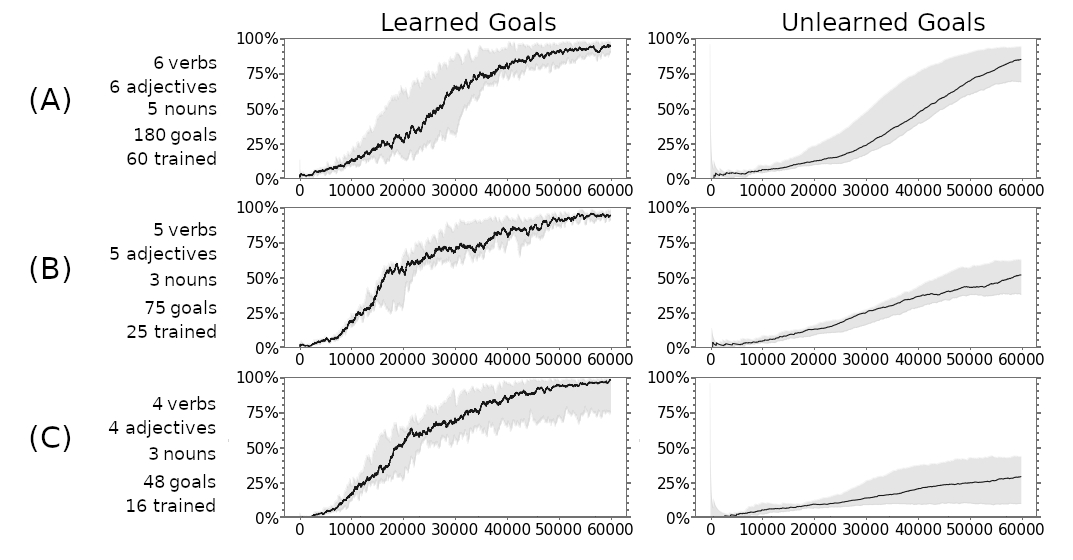}
    \caption{\textbf{Rolling success rates for learned and unlearned goals with different compositionality scales.} \textbf{(A)} Agents trained with all six verbs, all six adjectives, and all five nouns. \textbf{(B)} Agents trained with five verbs, five adjectives, and three nouns. \textbf{(C)} Agents trained with four verbs, four adjectives, and three nouns. Shaded areas represent 99\% confidence intervals for the ten agents.}
    \label{fig:generalization}
\end{figure}

Next, Fig. \ref{fig:generalization} \textbf{(A)} shows the success rate comparison between learned and unlearned goals under the \textit{all curiosity} condition for each action category. 
These plots show that the test performance for learned goals developed significantly faster than the case for the unlearned goals.
This indicates that actions are generated only for exactly learned compositional imperative sentences in the early phase, but the system generalizes to novel, unlearned ones in the later phase.

It is noted that the performance of the \textit{no curiosity} case is not completely saturated at 60,000 epochs (41 percent). Therefore, we conducted another simulation of this case up to 120,000 epochs to examine whether its performance could further improve. It appears that the performance of the \textit{all curiosity} case at 60,000 epochs (85.1 percent) is comparable to that of the \textit{no curiosity} case at 120,000 epochs (80.8 percent; see Figure \ref{fig:no_curiosity_120k}). This result indicates that the \textit{no curiosity} case can gain a generalization performance similar to the \textit{all curiosity} case, but its developmental speed is significantly slower than that of the \textit{all curiosity} case.

For comparison with the proposed model, we implemented a conventional baseline architecture based on the Soft Actor-Critic (SAC) algorithm \cite{pmlr-v80-haarnoja18b}, in which both the actor and critic networks are equipped with GRU recurrent layers \cite{recurrent}. Details of this baseline model are provided in the Experiment 1 section in Supplementary Material. 
Although Figures \ref{fig:baseline_results_six_tasks} and \ref{fig:baseline_results} show the best results achieved through various trials with different hyperparameter settings, its performance was substantially lower than that of the proposed model employing active inference for curiosity-driven exploration.


Video S1 in the Supplementary Material shows examples of the behaviors of robots with the \textit{all curiosity} level. 
It can be seen that in the intermediate phase of development, the robot often acts with play-like behavior without achieving specified goals. In the final phase of development, it quickly and accurately achieves its goals.

\subsection*{Further analysis}
\begin{figure}[ht!]
    \centering
    \includegraphics[width=.9\textwidth]{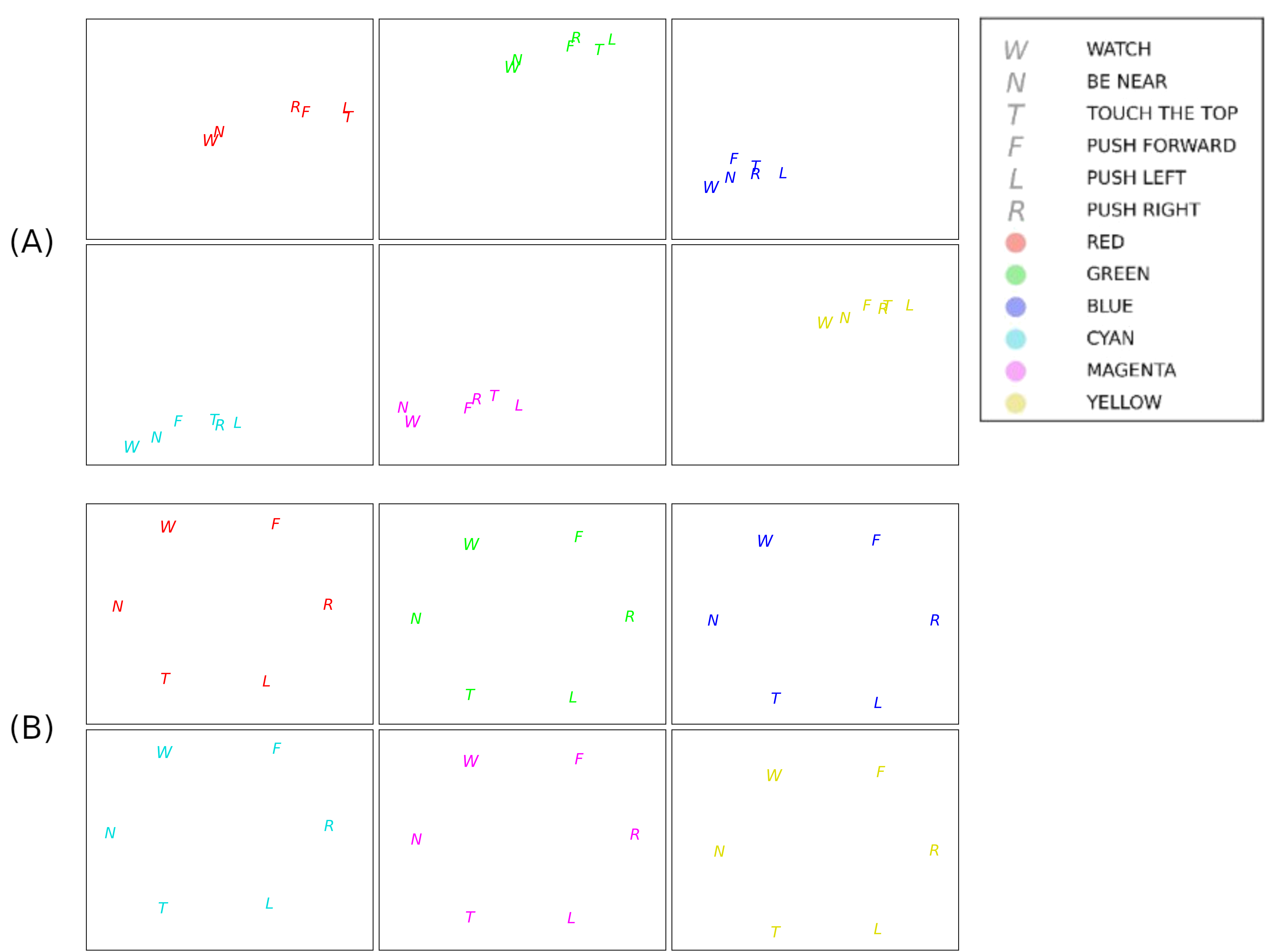}
    \caption{\textbf{PCA for language latent variables in the case of \textit{all curiosity.}} PCA of latent variables corresponding to the command voice. \textbf{(A)} Halfway through development (after 30,000 epochs). \textbf{(B)} complete development (after 60,000 epochs).}
    \label{fig:all_curiosity_composition}
\end{figure}

Post-hoc analyses were conducted to characterize the internal representations developed. Fig. \ref{fig:all_curiosity_composition} (A) and (B) show analysis of the internal representation by applying Principal Component Analysis (PCA) to the approximated posterior latent states corresponding to the command voice input, at 30,000 and 60,000 epochs in the development, respectively.
Each panel shows alignment of the principal component values among different verbs with the same adjective (color) averaged over all different object nouns.

In the case of full development (after 60,000 epochs) shown in Fig. \ref{fig:all_curiosity_composition} (B), it can be seen that the alignment of verbs repeats for all different colors.
This means that the internal representation for command sentences is factorized with different dimensions of verb and adjective/color after the full development.
On the other hand, in the case of halfway through development (after 30,000 epochs) shown in Fig. \ref{fig:all_curiosity_composition} (A), the alignments of verbs across different colors are partially broken, especially among ``touch the top'' (T), ``push left'' (L), and ``push right'' (R).
From these observations it can be said that the structure for systematicity compositionality can be developed gradually in the course of developmental learning.

\begin{figure}[ht!]
    \centering
    \includegraphics[width=\textwidth]{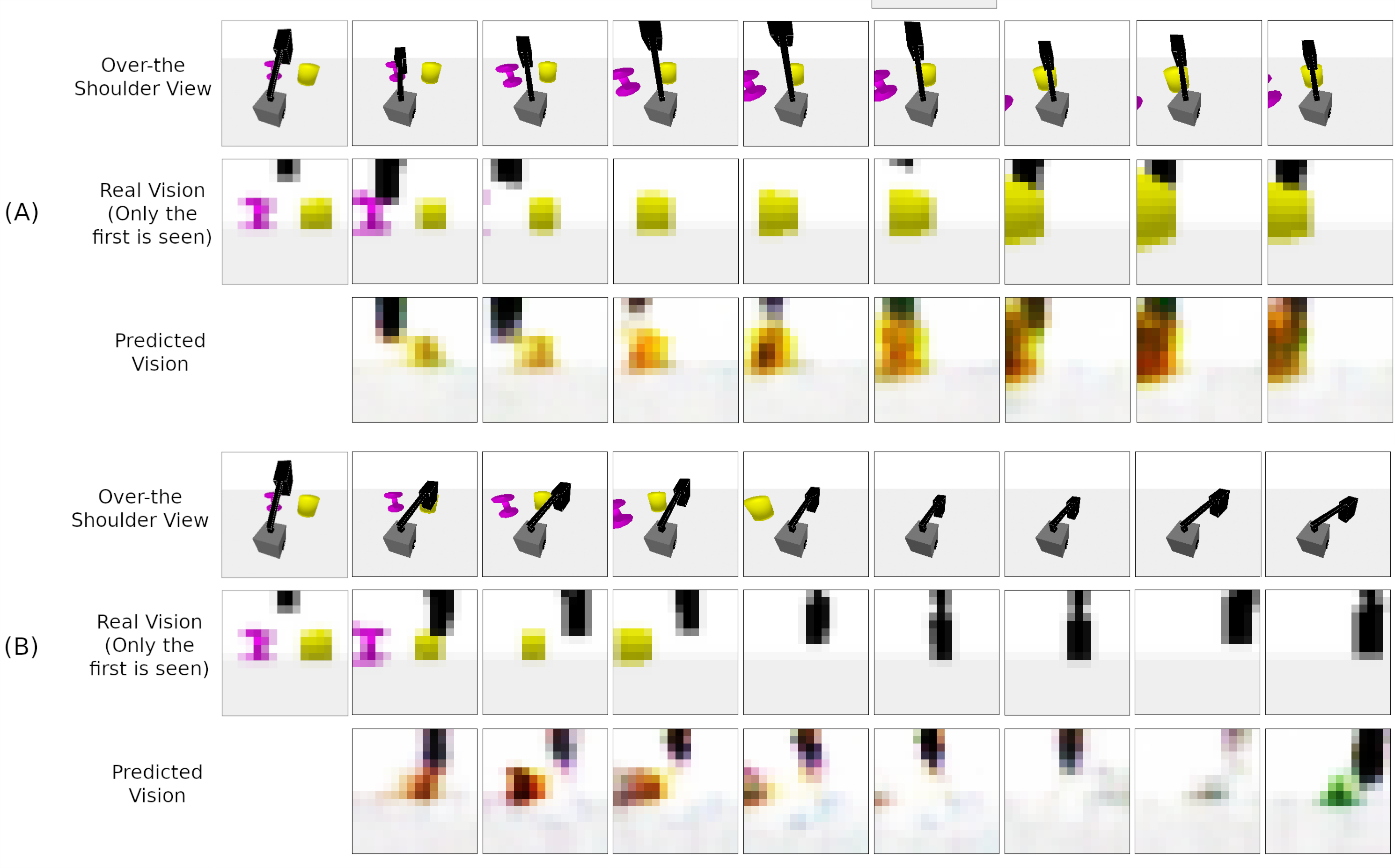}
    \caption{\textbf{Mental plans generated by the robot}. This robot is commanded to touch the top of the yellow pillar. The first row displays the ground truth from a view over the robot's right shoulder. The second row displays the visual sequence of the ground truth. The third row shows the visual sequence of mental planning. \textbf{(A)} The case for the end of complete developmental learning, and \textbf{(B)} for the case of halfway developed.}
    \label{fig:planning}
\end{figure}

In the current model, the robot's knowledge of the environment should become richer over the course of exploratory learning.
For confirmation of this idea, we examined the capability of the robots in generating mental plans for achieving goals without accessing the sensory inputs except the initial step for an episode trial as compared between the half-trained case and the fully trained case.
A robot's mental planning can be visualized by allowing it to receive real sensory observation only at the initial step, after which the robot must rely entirely on its own internal predictions. In this setting, the robot views its predicted sensory observations as if they are true inputs. This process may be likened to a dreamlike state or hallucinatory simulation, in which the robot mentally simulates future events based on its internal model of the world. 
Fig. \ref{fig:planning} \textbf{(A)} illustrates a simulation example for the fully trained case. 
The robot was commanded to touch the top of the yellow pillar. 
In that figure, the first row shows the ground truth environment from a view behind the robot's shoulder. 
The second row shows what the robot would truly observe if it were not in this planning setting. The third row shows the robot's look ahead predictions for visual observations, which it interprets as if they are real.
It can be seen that even with only the first step sensory observation, the robot could generate mostly accurate future look-ahead prediction for sensation as well as motor command.
These predictions are sufficiently accurate for the robot to maintain an internal conceptualization of the environment and complete its command in the case of the end of the developmental learning. 
Fig. \ref{fig:planning} \textbf{(B)} illustrates the same robot after only half of its training in the same scenario. In this case, the robot's predictions are inaccurate, causing it to wander and view an object which does not actually exist.

\subsection*{Effects of scale in compositions: Experiment 2}
This experiment examines the effects of scales of compositionality in learned examples to the generalization performance.
For this purpose, experiments were conducted using a reduced number of words for generating imperative sentences.
While the previous basic setup used sentences composed of 6 verbs, 6 adjectives, and 5 object nouns as the full scale case, the middle scale case was prepared with 5 verbs, 5 adjectives, and 3 object nouns, and the small scale case with 4 verbs, 4 adjectives, and 3 object nouns. The exact words used for each setup are listed in Table \ref{table:collections}.
For all scaling cases, only one third was used for learning examples while the remaining two thirds were used for generalization testing.
Other experimental conditions were the same as in Experiment 1.

The experimental results are shown in Fig. \ref{fig:generalization}. It can be seen that although the learned goal test cases show equally high performance for all scales of compositionality, the generalization testing for unlearned goal case shows that the success rate in the final trial decreases significantly (85.1 percent to 29 percent) as the compositionality scale decreases.
This indicates that the generalization performance severely depends on the scale of compositionality in learning examples.

\begin{table}[h]
\centering
\begin{tabular}{|c|c|c|c|}
\hline
Name & Verbs & Adjectives & Nouns \\
\hline\hline
Largest Vocabulary
& \makecell{Watch\\ Be Near\\ Touch the Top\\ Push Forward\\ Push Left\\ Push Right}
& \makecell{Red\\ Green\\ Blue\\ Cyan\\ Magenta\\ Yellow}
& \makecell{Pillar\\ Pole\\ Dumbbell\\ Cone\\ Hourglass} \\
\hline
Reduced Vocabulary
& \makecell{Watch\\ Be Near\\ Push Forward\\ Push Left\\ Push Right}
& \makecell{Red\\ Green\\ Blue\\ Cyan\\ Magenta}
& \makecell{Pillar\\ Pole\\ Dumbbell} \\
\hline
Smallest Vocabulary
& \makecell{Watch\\ Push Forward\\ Push Left\\ Push Right}
& \makecell{Red\\ Green\\ Blue\\ Cyan}
& \makecell{Pillar\\ Pole\\ Dumbbell} \\
\hline
\end{tabular}
\caption{\textbf{Words for training in three ways.} Verbs, adjectives, and nouns which are used for training agents in three different ways.}
\label{table:collections}
\end{table}

\subsection*{Exception rule handling: Experiment 3}

This experiment examined how robots can acquire exception-handling rules through developmental learning.
While most command–action mappings were preserved, the commands ``watch magenta pillar'' and ``be near green pole'' were swapped: success required performing the \textit{other} goal, not the one commanded. These mismatches required the robot to override its learned generalized knowledge.

This simulation experiment was conducted using the same model parameters used in the previous experiments with 10 robots with 60,000 epochs of developmental trials.
At the end of development the average success rate among 10 robots was 83 percent for the learned goals, 74 percent for unlearned goals, and 51 percent for the exception-handling cases.
The average success rate for the exception-handling cases is not so high (only some individuals are successful) which should be reasonable by considering the task complexity.

\begin{figure}[ht!]
    \centering
    \includegraphics[width=.55\textwidth]{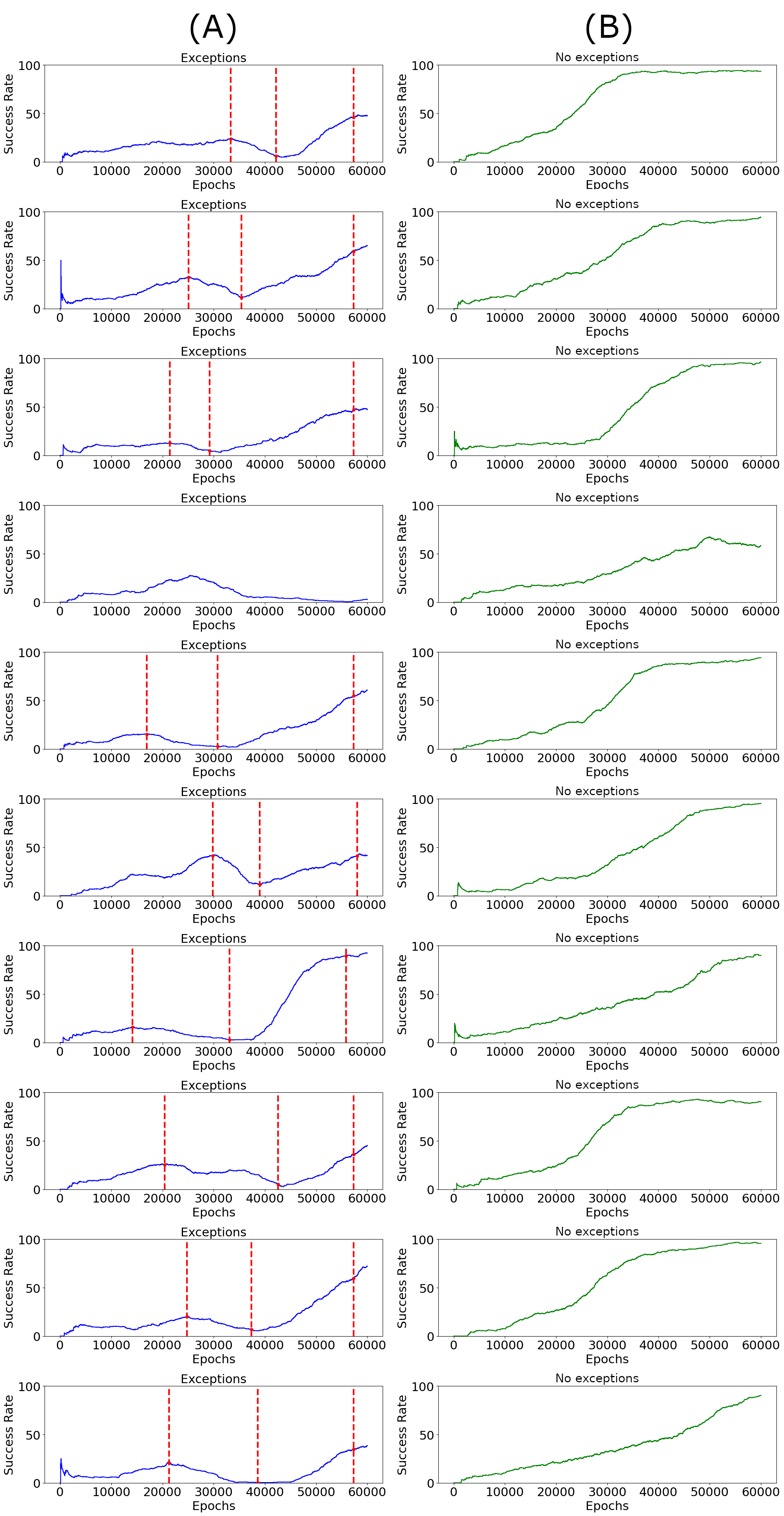}
    \caption{\textbf{Comparison of the performance curve with and without applying the exception handling rules for 10 individuals.} \textbf{(A)} The development of success rate in achieving two goals which are swapped as exceptions. Red vertical lines depict the peaks and valleys of the learning, as defined in supplementary text. \textbf{(B)} The development of success rate in achieving the same two goals without applying the exception handling rules.}
    \label{fig:exceptions}
\end{figure}

Panel \textbf{A} in Fig. \ref{fig:exceptions} shows the rolling success rate for achieving the exception goals for each of 10 individual robots while Panel \textbf{B} shows the success rate for the same goals but without applying the exception-handling rules.
The final success rate for the exception-handling case is diverse, ranging from 5 percent to 90 percent as can be seen in Panel \textbf{A}.
It was also discovered that 9 out of 10 robots trained with these exceptions exhibit characteristic U-shaped curves: early success, followed by a drop, and eventual recovery with higher success rate than the earlier one. 
In contrast, monotonic increase of success rate can be seen in all 10 individuals in the case of learning without the exception-handling rules.  
Statistical comparisons (detailed in the Supplementary Text) confirm that U-shaped patterns are significantly more prevalent in the exception condition than in the control ($p = .0001$). 

\begin{figure}[ht!]
    \centering
    \includegraphics[width=.7\textwidth]{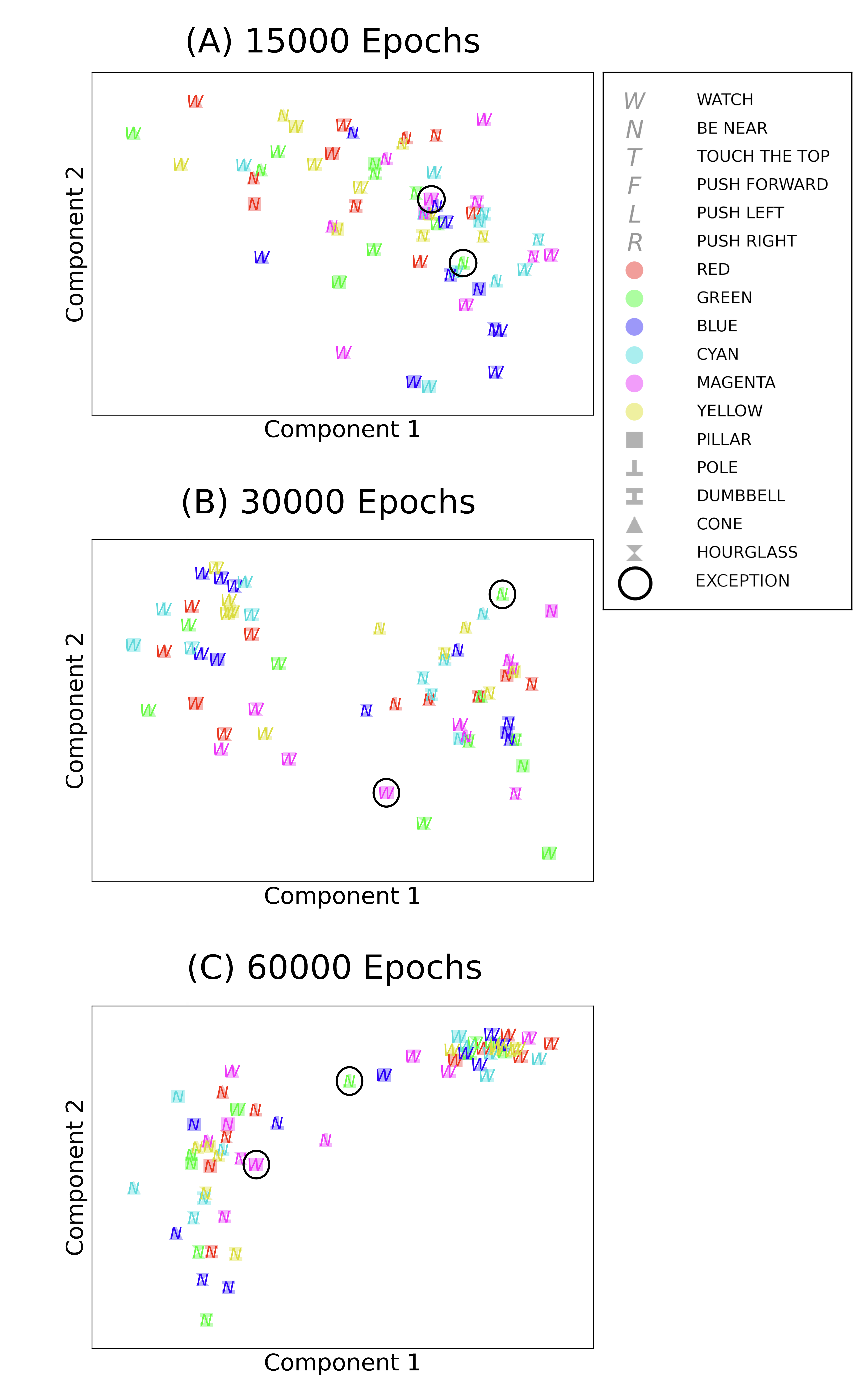}
    \caption{\textbf{PCA for language latent variables for an individual developed with the exception handling rules.} 
    \textbf{(A)} Plot at 15,000 epochs, 
    \textbf{(B)} plot at 30,000 epochs, 
    and \textbf{(C)} plot at 60,000 epochs wherein the circled ``N'' and circled ``W'' denote the sentences applied with the exception handling.}
    \label{fig:exceptions_compositions}
\end{figure}

In Fig.~\ref{fig:exceptions_compositions}, PCA illustrates how the internal representations of goals with the tasks ``watch'' or ``be near'' in the seventh robot in Fig. \ref{fig:exceptions} \textbf{(A)} evolve in a manner consistent with the U-shaped success rates. 
(Note that goals with other tasks are not shown for better visualization purpose.)
In panel~\textbf{A}, early in training, goal embeddings are muddled without clear structure, reflecting a learning phase with minimal generalization. 
In panel~\textbf{B}, midway in training, the exception command ``watch magenta pillar'' is embedded with other ``watch'' goals, while ``be near green pole'' clusters are embedded with other ``be near'' goals, despite these associations being incorrect. These observations indicate that overgeneralization has occurred.
In panel~\textbf{C}, late in training, ``watch magenta pillar'' is now embedded near ``be near'' goals, and ``be near green pole'' near ``watch'' goals, indicating that the robot has correctly handled these exceptions as swapped pairs.
Here, it can be said that the representational redescription took place in the course of developmental learning of exception-handling rules.

\section*{Discussion}

This study investigated how robots can develop action and language through self-exploration by amortizing active inference with deep learning. The experiments were designed to test five specific hypotheses, and the results provide clear support for each.  

\textbf{H1: Generalization is enhanced by compositional scale.}  
The experiments confirmed that larger vocabularies of verbs, adjectives, and nouns led to greater generalization. Robots trained with richer compositional repertoires achieved higher success rates on unlearned actions, whereas smaller vocabularies constrained generalization severely. Our previous study \cite{vijayaraghavan2021grounding} on supervised training of language and action for an arm robot also showed that compositional generalization improved as the size of verb-noun combinations in training increased. However, that work was limited in scale, examining only cases from $3 \times 3$ to $5 \times 8$ verb-noun combinations, where success rates for unlearned goals improved only from 57\% to 71\% under the condition of 80\% training. By contrast, the present study examined much broader scaling, ranging from 48 to 180 possible compositions, while using only 33\% of them for training. Under these conditions, generalization performance improved dramatically from near 25\% to 85\%. This contrast highlights that scale plays a critical role in enhancing generalization, and that curiosity-driven developmental learning provides a more powerful mechanism than supervised schemes under conditions of limited input. The finding also connects to the classical ``poverty of the stimulus'' problem raised by Chomsky \cite{chomsky1980rules}, as it shows how compositionality enables powerful generalization from sparse training data. 
Once, we hypothesized that necessary training size could be proportional to summation of number of words appeared for each dimension instead of multiplication of it for all dimensions if compositionality size increases \cite{vijayaraghavan2021grounding}.
This hypothesis becomes more plausible given the results in this current study, which should be confirmed in much more scaled experiments in the future.

\textbf{H2: Curiosity combined with motor entropy enhances developmental learning.}  
These results support H2, indicating that curiosity-driven exploration facilitates the development of structured internal representations that enable compositional generalization under conditions of sparse experience. In particular, agents equipped with intrinsic motivation acquire such representations more efficiently and with fewer training epochs compared to those without curiosity. However, additional experiments extending the training of the no-curiosity condition to 120K epochs revealed that comparable levels of compositional generalization can eventually be achieved given sufficiently prolonged experience. This suggests that while curiosity is not strictly necessary for the emergence of structured representations, it plays a critical role in accelerating learning and improving sample efficiency. Therefore, the primary contribution of curiosity-driven exploration in the present framework lies in facilitating faster and more efficient acquisition of compositional structure rather than being the sole mechanism enabling it.
 
\textbf{H3: Generalization follows rote learning.}  
The results again align with this hypothesis: in early phases, the robot succeeds only on exactly learned sentence–action pairs, and it is only over time that the robot begins to generalize to novel combinations of known elements. 
This mirrors developmental patterns in infants, who often begin with rigid pairings before achieving broader generalization \cite{tomasello2003constructing}. 
Tomasello’s ``verb-island'' hypothesis, for example, emphasizes that children initially acquire verbs in isolated contexts before generalizing across broader structures \cite{tomasello2003constructing}, in the same way that this robot acquired trained goals first before generalizing to untrained goals.
Gerken \& Knight \cite{gerken2015infants} demonstrated that 10- to 11-month-old infants can generalize from just four linguistic examples under favorable conditions. 
Moreover, Gerken et al. \cite{gerken2014surprise} provide evidence that infants may generalize even from a single surprising example, suggesting that hypothesis-driven generalization can follow minimal exposure. These studies lend developmental credence to our observed progression from rote mapping toward flexible compositional generalization. 

\textbf{H4: Primitive actions precede complex actions.}  
Regarding the fourth hypothesis, “primitive actions precede complex actions,” our results provide only partial support. As shown in Fig. \ref{fig:evaluation} in the all curiosity condition, some relatively complex actions, such as "push left" and "push right", emerged more slowly than other behaviors, which is broadly consistent with the hypothesis. However, among the remaining actions, namely "watch", "be near", "push forward", and "touch the top", no clear or statistically significant differences were observed in their developmental speeds, despite differences in their presumed levels of action complexity.
This pattern differs somewhat from the original expectation of a monotonic relationship between action complexity and acquisition speed. One possible interpretation is that factors other than structural complexity, such as environmental affordance, reward structure, or perceptual accessibility, may play an important role in shaping the developmental trajectory of actions. 

\textbf{H5: Exception-handling rules exhibit U-shaped development.}    
The results from Experiment~3 provide strong support for this hypothesis. 
When robots were trained with two swapped command–action mappings, U-shaped performance trajectories were observed more frequently than robots trained without these exceptions.
Our analysis of the latent variables in the model network showed that overgeneralization takes place in the middle of development and such internal representation is redescribed to accommodate the exception-handling rules later.
This non-monotonic, U-shaped performance trajectory mirrors a well-established phenomenon in developmental psychology, in which children first succeed on irregular forms, later overgeneralize newly learned rules (e.g., producing ``goed'' for ``went''), and ultimately reorganize their internal representations to master both rules and exceptions. 
Classic accounts interpret these dynamics as evidence for \emph{representational redescription}, a restructuring of internal knowledge that enables more abstract, generative representations \cite{karmiloffsmith1992beyond}.  

Computational modeling has long shown that such U-shaped learning can emerge naturally from error-driven or distributed representations, including Rumelhart and McClelland’s connectionist model of English past-tense acquisition \cite{rumelhart1986pasttense}, the multilayer perceptron models of Plunkett and Marchman \cite{plunkett1991ushape, marchman1993acquisition}, and broader frameworks in computational developmental psychology \cite{mareschal2001computational}. Our robot simulations demonstrate that an analogous process arises in curiosity-driven active inference: the agent first relies on rote pairings, then applies generalized compositional mappings that overwrite earlier exceptions, and finally reconstructs its latent representation to encode the exception rules correctly.  

The parallels with infant development, including rote-to-generalization progression, prerequisite learning, the role of vocabulary scale, and representation redescription, suggest that the mechanisms implemented here capture some essential aspects of developmental psychology. 
More broadly, these results strengthen the view that reconstructing developmental processes in robots can offer insights into the ``poverty of the stimulus'' problem, showing how powerful generalization can arise from limited input when guided by intrinsic motivation, structured experience, and the principles of predictive coding and active inference \cite{asada2001cognitive, sugita2005cross, vijayaraghavan2021grounding, cangelosi2015developmental}.  

While the present model demonstrates the emergence of compositional generalization under curiosity-driven self-exploration, several limitations should be noted. First, although comparison with a Soft Actor-Critic baseline suggests that the proposed architecture benefits from the inclusion of a forward model for structuring latent representations, the current study is not intended to establish performance superiority over alternative AI models, but rather to elucidate qualitative developmental characteristics under varying conditions such as compositional scale and intrinsic motivation. In addition, the learning process is relatively slow, requiring on the order of 60,000 trials to achieve stable performance, indicating limited efficiency under the present setting.
 
Second, the present framework provides only a simplified abstraction of human development. Unlike human infants, whose language acquisition is preceded by rich sensorimotor and social experience and supported by caregiver scaffolding and developmental staging, our model isolates a minimal setting to investigate how compositional generalization and exception handling can emerge from intrinsically motivated exploration. Accordingly, the model should be understood as a controlled platform for examining core mechanisms rather than a comprehensive account of human language development.

These limitations naturally point to several important directions for future research. One key extension is to move beyond the current one-directional communication paradigm between tutors and robots. In the present study, learning is guided solely by externally provided commands and feedback, whereas human development is characterized by interactive and bidirectional communication \cite{lipschits2024integrative,suarezrivera2026pathways}. Future work should therefore incorporate mechanisms by which robots can actively solicit guidance, request clarification, or negotiate task difficulty, enabling more adaptive and socially grounded learning through interaction with caregivers.

Another important direction concerns improving learning efficiency by introducing structured developmental progression. The current model assumes learning from scratch without prior knowledge, whereas human learners acquire perceptual categories and basic motor repertoires before engaging in more complex compositional tasks. Incorporating pre-trained perceptual and motor primitives, together with curriculum learning in which simpler tasks are acquired prior to more complex combinations \cite{tomasello2003constructing,Clark2009}, may substantially accelerate learning and improve generalization. More broadly, integrating both staged learning and social interaction will be essential for bridging the gap between the present minimal framework and the richness of human developmental processes.

Another promising future direction concerns the development of robot-robot communication through the evolution of language. Previous research has explored this possibility from different perspectives: Steels introduced the framework of ``language games'' to study the emergence of shared vocabularies among agents \cite{steels1995self}, Miikkulainen and colleagues investigated the evolution of artificial language through evolutionary reinforcement learning \cite{li2006evolving}, and Taniguchi proposed the emergence of symbols using a collective predictive coding approach \cite{taniguchi2019symbol}. 
While these studies have demonstrated the possibility of emergent communication in an impressive manner, they still remain limited in that they only considered the emergence of object labeling or naming, whereas the evolution of action-related language, such as verbs, has been much less explored. 
In this context, the current study based on active inference could be extended to address the evolution of dynamic linguistic structures, including verbs. Since our model implements active inference within a variational recurrent neural network, it is naturally suited for capturing temporal and dynamic aspects of action and language. 
A future extension of this work toward multi-robot interaction under the framework of ``collective active inference'' or ``federated active inference'' may thus provide novel insights into the evolution of embodied language, moving beyond static object labeling toward dynamic and action-oriented communication.

\section*{Materials and Methods}

In this section, we present the model architecture employed in this study. 
The current model, as well as our earlier work \cite{tinker2024intrinsic}, extends a study by Kawahara et al. \cite{kawahara}.
That study demonstrated that curiosity-driven reinforcement learning can be achieved within the framework of active inference (AIF) \cite{parr_friston_2019, friston2016active}, in which motor behavior is reinforced in the direction which minimizes expected free energy. 
More details are shown in the ``Free energy principle, Active Inference, and Kawahara Model'' section of the Supplementary Materials, along with a brief introduction of the free energy principle (FEP) and AIF.

\subsection*{The Employed Model}
The current model, as well as our previous one \cite{tinker2024intrinsic}, extends the approach proposed by Kawahara et al. \cite{kawahara} by implementing both the forward model and actor-critic using a variational recurrent neural network (VRNN) \cite{chung2015recurrent} in order to deal with temporal complexity and stochasticity inherent in robot–environment interactions.

The expected free energy $G$ can be computed as:
\begin{align} \label{eq:G_method}
G_t &=  -\underbrace{\eta D_{KL}[q(z_{t}|o_{t}, a_{t-1}, h_{t-1})||p(z_{t}|a_{t-1}, h_{t-1})]}_{\text{Curiosity}} - \underbrace{r(s_t, a_t)}_{\text{Extrinsic Reward}} - \underbrace{\alpha \mathcal{H}(\pi_\phi(a_t|h_{t}))}_{\text{Entropy}} 
\end{align}
This equation is derived by replacing $w$, the probabilistic model learning parameter used in Eq.~\ref{eq:G_1}, with $z$, the probabilistic model state.
The weighting coefficients $\eta$ and $\alpha$ are introduced to scale the contributions of the curiosity and motor entropy terms, respectively.
%
%
The complexity term is computed as Kullback–Leibler divergence (KLD) between the approximated posterior distribution and the prior distribution over the latent variables at each time step.
Both distributions are modeled as Gaussian distribution with time-dependent means and standard deviations.
The approximated posterior is conditioned on the current sensory observation and the previous latent control variable, while the prior is conditioned only on the previous latent control variable. The resulting KLD thus reflects the information gain from that sensory observation, which is driven by the motor command executed at the previous time step.
Therefore, exploration of more novel situations (i.e., curiosity-driven exploration) tends to result in higher information gain through larger complexity.
The motor entropy in the third term of Eq.~\ref{eq:G_method} reflects the expected uncertainty of the policy, and is computed as the negative expected log-probability of generating a motor command $a_t$ conditioned on the latent control variable $h_{t}$.

By adopting an analogous approach to the Kawahara model, the policy for generating a motor command $a_t$ is trained to minimize the expected free energy $G_t$ (Eq.~\ref{eq:G_method}) through RL using the Soft Actor-Critic (SAC) algorithm \cite{pmlr-v80-haarnoja18b}.
Accordingly, the $Q_t$ value is updated as:
\begin{align}\label{eq:bellman_plus_curiosity_method}
Q_t &= r_t + \eta D_{KL}[q(z_{t}|o_{t},h_{t-1})||p(z_{t}|h_{t-1})] + \alpha \mathcal{H}(\pi_{\phi} (a_{t+1} | h_{t}))\nonumber \\ 
+ &\gamma (1 - done_t) \mathbb{E}_{o_{t+1} \sim D, a_{t+1} \sim \pi_\phi} [Q_{\bar{\theta}}(o_{t+1}, a_{t+1})].
\end{align}
\noindent
The first term $r_t$ represents the extrinsic reward. The second term $D_{KL}[q(z_{t}|o_{t},h_{t-1})||p(z_{t}|h_{t-1})]$ is the intrinsic reward for curiosity, scaled by a positive coefficient $\eta$. The third term $\mathcal{H}(\pi_{\phi} (a_{t+1} | h_{t}))$ is the intrinsic reward for motor entropy, scaled by a positive coefficient $\alpha$. The fourth term is the bootstrapped estimate of the next step's value, $\widehat{Q_{t+1}}$, which is weighted by a discount rate parameter $\gamma \in [0,1]$.
The variable $done_t$ is zero for all steps except the episode's final step, where it is set to one. This restrains the definition of $Q_t$ to steps within the episode.
The critic $Q_{\theta}(o_{t+1}, a_{t+1})$ is trained to generate $\widehat{Q_t}$, approximation of $Q_t$. The target critic $Q_{\bar{\theta}}(o_{t+1}, a_{t+1})$ is maintained for stability in the critic's training. Initially identical to the critic, the target critic is updated via Polyak averaging such that $\bar{\theta} \leftarrow \tau\theta + (1 - \tau) \bar{\theta}$ with $\tau \in [0,1]$. The actor $\pi_\phi(o_t)$ is trained to generate motor commands $a_t$ which maximize the critic's predictions of value. To mitigate positive bias, it is common to train multiple separate critics (each with its own target critic) \cite{pmlr-v80-haarnoja18b}. The actor is trained using the minimum predicted value across critics. Our model employs two separate critics.



The forward model is trained dynamically over the course of exploratory learning by optimizing the model parameters $\psi$ to minimize the evidence free energy $F_{\psi}$ (Eq. \ref{eq:F}) after each trial episode. 
The exact implementation of this process is described in the supplementary material subsection, Details of the Model Architecture.

\subsubsection*{Robot Actions}
\label{sec:Def_Actions}

The robot and the objects were simulated in PyBullet, the Python physics simulator. 
Each wheel’s velocity was bounded within the range of $[-10, 10]$ meters per second. For scale, the robot’s body is a cube measuring 2 meters along each dimension (length, width, and height).
The robot's arm features two joints: yaw, which rotates left or right within a range of $[-30^\degree, 30^\degree]$, and pitch, which rotates forward or upward within a range of $[0^\degree, 90^\degree].$
For smooth movement, the robot's wheel and arm velocities were implemented with linear interpolation from the current to the target velocities. 

We defined success criteria for each action category, which determined whether or not the robot earned an extrinsic reward by completing a goal. 
The distance between the robot and an object was measured from the object's center to the center of the robot's body. The robot was considered to be ``facing the object'' when the angular deviation between the robot's forward direction and the line connecting it to the object was less than 15 degrees.

\textbf{Watch:} The robot faces the object between 6 and 10 meters of distance. This must be maintained for 6 steps in a row.

\textbf{Be Near:} The robot faces the object with distance of less than 6 meters, without touching the object. This must be maintained for 5 steps in a row.

\textbf{Touch the Top:} The robot's hand contacts with the object while the center of the hand is at least 3.75 meters above the floor. This must be maintained for 3 steps in a row.

\textbf{Push Forward:} The robot pushes the object farther than .1 meters with respect to the robot's facing direction. This must be maintained for 3 steps in a row.

\textbf{Push Left:} The robot pushes the object to the robot's left farther than .2 meters while the robot's wheels have velocities below 5 meters per second (requiring use of the arm). This must be maintained for 3 steps in a row.

\textbf{Push Right:} Same as \textbf{Push Left,} but in the opposite direction.

There are constraints in rewarding for actions which are described in the ``Constraints in Performing Actions'' subsection in the Supplementary Materials.


\clearpage 

%
\bibliography{bib.bib} 
\bibliographystyle{sciencemag}

%
%
%
%
%
%


\section*{Acknowledgments}
Thank you to the Okinawa Institute of Science and Technology (OIST) and colleagues in OIST's Cognitive Neurorobotics Unit and Neural Computation Unit for supporting this work. We also thank Taro Toyoizumi for valuable discussions on acquiring exception-handling rules.

\paragraph*{Funding:}
T. J. T. was funded by OIST graduate school. 
J. T. and K. D. were supported by OIST research funding. This work was funded by the Japan Society for the Promotion of Science (JSPS) KAKENHI, Transformative Research Area (A): unified theory of prediction and action 26H01186 to J. T. and 23H04975 to K. D.

\paragraph*{Author contributions}
T.J.T. and J.T. designed the model and simulation. T.J.T., K.D., and J.T. designed the experiments. T.J.T. performed all experiments and data analysis. T.J.T. wrote the paper. K.D. and J.T. edited the paper. J.T. supervised the study.

\paragraph*{Competing interests:}
Authors declare that they have no competing interests.

\paragraph*{Data and materials availability:} All files related to this project are available from a repository at the Okinawa Institute of Science and Technology. Files are also available through DOI 10.5061/dryad.3n5tb2s01, ``Robotic agents for `Curiosity-Driven Development of Language.' ''

The code used in the experiments is accessible via \url{https://github.com/oist-cnru/Curiosity-Driven-Development}.

The authors are grateful for the help and support provided by colleagues in OIST's Cognitive Neurorobotics Research Unit and Neural Computation Unit. 
\newpage


\renewcommand{\thefigure}{S\arabic{figure}}
\renewcommand{\thetable}{S\arabic{table}}
\renewcommand{\theequation}{S\arabic{equation}}
\renewcommand{\thepage}{S\arabic{page}}
\setcounter{figure}{0}
\setcounter{table}{0}
\setcounter{equation}{0}
\setcounter{page}{1} 


\begin{center}
\section*{Supplementary Materials for\\ \scititle}

Theodore Jerome Tinker$^1$,
Kenji Doya$^1$,
Jun Tani$^1$\\ 
\small$^1$ Okinawa Institute of Science and Technology, Okinawa, Japan.\\
\text{*} To whom correspondence should be addressed; E-mail: jun.tani@oist.jp.
\end{center}

\subsubsection*{This PDF file includes:}
Supplementary Text\\
Figs. S1 to S7\\
Tables S1 to S8\\

\subsubsection*{Other Supplementary Materials for this manuscript:}
Video S1: \url{http://youtube.com/watch?v=RJd-6mkW-bM}

\noindent
\textbf{Video mentioned in the results section.} 

\newpage



\subsection*{Supplementary Text}

For future reference, table \ref{table:variables} includes definitions for relevant variables.

\movie[example]{Training example.}
{Compares a robot with \textit{all curiosity} mid-training and after training.}

\begin{table}[]
\centering

\begin{multicols}{2} 

\begin{tabular}{|>{\raggedright}p{.8in}|>{\raggedright\arraybackslash}p{2in}|}
\hline
Variable & Definition \\
\hline
$o_t$ & Observation at time $t$ \\
$o_{t,i}$ & $i^{th}$ part of observation $o_t$ \\
$o_{t,v}$ & Our agent’s $o_{t,0}$, vision \\
$o_{t,ta}$ & $o_{t,1}$, touch \\
$o_{t,p}$ & $o_{t,2}$, proprioception \\
$o_{t,cw}$ & $o_{t,3}$, command voice \\
$o_{t,fw}$ & $o_{t,4}$, feedback voice \\
$a_t$ & Motor Command \\
$r_t$ & Extrinsic reward \\
$done_t$ & Final step of episode \\
$mask_t$ & Steps inside episode \\
$R$ & Recurrent replay buffer\\
$\pi$ & Actor \\
$\phi$ & Actor’s parameters \\
$Q$ & Critic \\
$\theta$ & Critic's parameter \\
$\bar{\theta}$ & Target critic's parameter \\
$\tau$ & Critic's soft update coefficient \\
\hline
\end{tabular}

\vspace{1em}

\begin{tabular}{|>{\raggedright}p{.8in}|>{\raggedright\arraybackslash}p{2in}|}
\hline
Variable & Definition \\
\hline
$f$ & Forward model \\
$\psi$ & Forward model parameters \\
$\gamma$ & Discount for future rewards \\
$\alpha$ & Importance of motor entropy \\
$\eta$ & Importance of curiosity \\
$\eta_i$ & $\eta$ for $i^{th}$ part of observation \\
$p(z_t)$, $q(z_t)$ & Prior, approximated posterior \\
$\mu$, $\sigma$ & Mean, standard deviation \\
$h_t$ & RNN hidden state \\
$z_t$ & Sample from posterior \\
$enc_i$ & Encoder for $o_{t,i}$ \\
$\psi^{enc}_i$ & $f$ parameters for $enc_i$ \\
$dec_i$ & Decoder for $o_{t,i}$ \\
$\psi^{dec}_i$ & $f$ parameters for $dec_i$ \\
$MLP^{prior}_i$ & Multilayer for prior for $o_{t,i}$ \\
$MLP^{post}_i$ & Multilayer for approximated posterior for $o_{t,i}$ \\

\hline
\end{tabular}

\end{multicols}
\caption{\textbf{Definitions of variables.}}
\label{table:variables}
\end{table}

\subsection*{Free Energy Principle, Active Inference, and Kawahara Model}
\label{sec:suppl_kawahara}

We begin by describing predictive coding and active inference (AIF), which are grounded in the free energy principle (FEP) \cite{friston2005}. The FEP posits that biological and artificial agents maintain their existence by minimizing variational free energy, which is an upper bound on sensory surprise. 
In perception, this process is often instantiated as predictive coding \cite{friston2005, rao1999predictive, hohwy2013predictive, clark2015surfing}, wherein internal models reconstruct sensory inputs by updating beliefs or latent variables by minimizing the reconstruction errors. More formally, this is minimizing evidence free energy defined for past observations. 
In motor command generation, the FEP framework extends to AIF \cite{friston2011, parr_friston_2019}, where agents minimize the future prediction error (quantified as expected free energy) by optimizing the latent variables and motor commands in the future.
These two processes are tightly coupled and must be considered jointly in embodied cognition systems.

We next introduce the work of Kawahara et al.~\cite{kawahara}, who proposed a novel reinforcement learning (RL) scheme that integrates (AIF).

In the Bayesian framework, the true posterior probability distribution $p(z_t|o_t)$ over latent variables $z_t$, conditioned on sensory observations $o_t$, is given by Bayes' rule:

$$p(z_t|o_t) = \frac{p(o_t|z_t)p(z_t)}{\int{p(o_t, z_t) dz}}$$

\noindent
Here, $p(z_t)$ denotes the prior. The denominator, called the evidence, is usually intractable; to overcome this, variational Bayes introduces an approximation of the posterior $q(z_t)$. This is optimized to minimize the Kullback-Leibler divergence (KLD) between the approximated posterior $q(z_t)$ and the true posterior $p(z_t|o_t)$.

\begin{align}
D_{KL} [q(z_t) || p(z_t|o_t)] &= \int q(z_t)\log\frac{q(z_t)}{p(z_t|o_t)} dz_t \nonumber \\
&= \int q(z_t)\log\frac{q(z_t)p(o_t)}{p(z_t,o_t)} dz_t \nonumber \\
&= \int q(z_t)\log\frac{q(z_t)p(o_t)}{p(z_t)p(o_t|z_t)} dz_t \\
&= F + \log{p(o_t)}
\end{align}

\noindent
The term $F$ here is the evidence free energy, equal to 
\begin{align} \label{eq:F} 
F_t = \underbrace{D_{KL}[q(z_t)||p(z_t)]}_{\text{Complexity}} - \underbrace{\mathbb{E}_{q(z_t)}[\log p(o_{t+1}|z_t)]}_{\text{Accuracy}}.
\end{align}
\noindent
Since $p(o_t)$ is constant for a given sensory observation, minimizing KLD is equivalent to minimizing $F_t$.
Therefore, the optimal posterior approximation is:
\begin{align} \label{eq:Farg_min_F} 
q^*(z_t) = \underset{q(z_t)}{\text{arg min }} F_t
\end{align}

In active inference, the agent minimizes expected free energy $G_\tau$ at a future time step $\tau \geq t+1$. This is the expected value of the evidence free energy under the predictive distribution of future outcomes \cite{kawahara}.

\begin{align}\label{eq:this_one} 
G_\tau &= \mathbb{E}_{p(o_\tau | z_\tau)} [F] \nonumber \\
&= \mathbb{E}_{p(o_\tau | z_\tau)} [\int q(z_\tau) \log \frac{q(z_\tau)}{p(o_\tau,z_\tau)} dz] \nonumber \\
&= \mathbb{E}_{p(o_\tau|z_\tau)} [\mathbb{E}_{q(z_\tau)} [\log \frac{q(z_\tau)}{p(z_\tau|o_\tau)} - \log p(o_\tau)]].
\end{align}

\noindent 
Recalling that $q(z_\tau|o_\tau) q(o_\tau) = q(o_\tau,z_\tau)$, we approximate:

\begin{align} 
G_\tau &\approx \mathbb{E}_{q(o_\tau,z_\tau)} [\log \frac{q(z_\tau)}{q(z_\tau|o_\tau)} - \log p(o_\tau)] \nonumber \\
&= - \mathbb{E}_{q(o_\tau, z_\tau)} [\log \frac{q(z_\tau|o_\tau)}{q(z_\tau)}] - \mathbb{E}_{q(o_\tau)} [\log p(o_\tau)] \nonumber \\
&= - \underbrace{\mathbb{E}_{q(o_\tau)} [\overbrace{D_{KL} [q(z_\tau|o_\tau) || q(z_\tau)]}^{\text{Bayesian Surprise}}]}_{\text{Epistemic Value or Mutual Information}} - \underbrace{\mathbb{E}_{q(o_\tau)} [\log p(o_\tau)]}_{\text{Extrinsic Value}}. 
\end{align}

\noindent
The first term, $I(z_\tau,o_\tau) = \mathbb{E}_{q(o_\tau)}[D_{KL}[q(z_\tau|o_\tau)||q(z_\tau)]]$, is the mutual information (or Bayesian surprise). This depicts expected information gain based on new sensory observation $o_\tau$, and can be expressed as:
$$I(z_\tau,o_\tau) = \underbrace{H(z_\tau)}_{\text{Shannon Entropy}} - \underbrace{H(z_\tau|o_\tau)}_{\text{Conditional Entropy}}.$$
\noindent
The second term, $p(o_\tau)$, represents log-likelihood of the preferred sensory observation. This is specified as the extrinsic reward designed by the experimenters.
For the intrinsic value to reflect mutual information or information gain, and the extrinsic value to reflect expected free energy, is the same as the way shown by Friston's group in the study of active inference \cite{parr_friston_2019, friston2016active}.
Separating $o_t$ into $o_t$ and $a_t$, we rewrite the expected free energy as: 

\begin{align} \label{eq:applying_D} 
G_\tau &= -\mathbb{E}_{q(o_\tau, a_\tau, z_\tau)} [\log \frac{p(z_\tau|o_\tau,a_\tau)}{q(z_\tau)}] - \mathbb{E}_{q(o_\tau, a_\tau)} [\log p(o_\tau, a_\tau)] \nonumber \\
&= -\mathbb{E}_{q(o_\tau, a_\tau, z_\tau)} [\log \frac{p(z_\tau,a_\tau|o_\tau)}{q(z_\tau)p(a_\tau|o_\tau)}] - \mathbb{E}_{q(o_\tau, a_\tau)} [\log p(o_\tau, a_\tau)] \nonumber \\
&\approx -\mathbb{E}_{q(o_\tau, a_\tau, z_\tau)} [\log \frac{q(z_\tau|o_\tau)q(a_\tau|o_\tau,z_\tau)}{q(z_\tau)p(a_\tau|o_\tau)}] - \mathbb{E}_{q(o_\tau, a_\tau)} [\log p(o_\tau, a_\tau)] \nonumber \\
&= -\mathbb{E}_{q(a_\tau | o_\tau, z_\tau)q(o_\tau)} [D_{KL}[q(z_\tau|o_\tau)||q(z_\tau)]] \nonumber \\
&- \mathbb{E}_{q(o_\tau,z_\tau)} [D_{KL}[q(a_\tau|o_\tau,z_\tau)||p(a_\tau|o_\tau)]] \nonumber \\
&- \mathbb{E}_{q(o_\tau,a_\tau)}[\log p(o_\tau,a_\tau)]. 
\end{align}

Kawahara et al.~\cite{kawahara} developed a forward model $f_w(o_\tau, a_\tau) \rightarrow \widehat{o}_{\tau+1}$ which learns to predict the future sensory observation $o_{\tau+1}$ based on $o_\tau$ and $a_\tau$ using a Bayesian Neural Network (BNN) \cite{bnn}. In this type of model, the network parameters $w_\tau$ are treated as random variables defined with Gaussian distribution. These parameters serve as latent causes of observed sensory transitions and can be interpreted as random latent variables for the generative model.
Therefore, $w_\tau$ corresponds to $z_\tau$.

Let the approximate posterior be defined as $q_\psi = \mathcal{N}(w_\tau|\mu,\sigma)$, with parameters $\psi = \{\mu,\sigma\}$. 
In this setting, the actor $\pi_\phi$ of a SAC can be trained to approximate $\pi_\phi(a_\tau|o_\tau) \approx q(a_\tau|o_\tau,w_\tau)$. This allows rewriting the expected free energy as:

\begin{align} \label{eq:replaced_X} 
G(o_\tau, a_\tau) &= -\mathbb{E}_{q(a_\tau | o_\tau, w_\tau)q(o_\tau)} [D_{KL}[q(w_\tau|o_\tau)||q(w_\tau)]] \nonumber \\
&- \mathbb{E}_{q(o_\tau,a_\tau)} [D_{KL}[\pi_{\phi}(a_\tau|o_\tau)||p(a_\tau|o_\tau)]] \nonumber \\
&- \mathbb{E}_{q(o_\tau,a_\tau)}[\log p(o_\tau,a_\tau)].
\end{align}

\noindent
Let us interpret the prior preference $\log p(o_\tau, a_\tau)$ as the extrinsic reward $r(s_\tau, a_\tau)$, where $s_\tau$ is the true environmental state. 
Bring focus to the current time step by setting $\tau = t$.
Because the forward model trains to predict $o_{t+1}$, we can further rewrite the expected free energy as:

\begin{align}\label{eq:G_1} 
G(o_t, a_t) &= -D_{KL}[q_\psi(w_t|o_{t+1})||q_\psi(w_t)] - \log p(o_t,a_t) \nonumber \\
&- D_{KL}[\pi_\phi(a_t|o_t)||p(a_t|o_t)] \nonumber \\
&= -D_{KL}[q_\psi(w_t|o_{t+1})||q_\psi(w_t)] - \log p(o_t,a_t) \nonumber \\
&- \int \pi_\phi(a_t|o_t) \log \pi_\phi(a_t|o_t) da_t + \int \pi_\phi(a_t|o_t) \log p(a_t|o_t) da_t\nonumber \\
&= -\underbrace{D_{KL}[q_\psi(w_t|o_{t+1})||q_\psi(w_t)]}_{\text{Curiosity}} - \underbrace{r(s_t, a_t)}_{\text{Extrinsic Reward}} - \underbrace{\mathcal{H}(\pi_\phi(a_t|o_t))}_{\text{Entropy}} - \underbrace{\mathbb{E}_{\pi_\phi(a_t|o_t)} [\log p(a_t^\ast|o_t)]}_{\text{Imitation}}
\end{align}

\noindent
Because $w_\tau$ represents the robot's probabilistic knowledge of its environment, the first term of Eq.~\ref{eq:G_1} can be said to represent the robot's gain in knowledge based on information acquired in a new sensory observation. 

In summary, the forward model is trained to minimize the evidence free energy $F$ (Eq.~\ref{eq:F}) by accurately reconstructing sensory observations and minimizing posterior complexity based on past experiences. 
Meanwhile, the actor-critic pair is trained to minimize expected free energy $G$, which includes an inverted complexity term (i.e., curiosity) and motor entropy to encourage exploration.
This leads to emergent tension in an adversarial relationship: the actor is encouraged to maximize information gain by increasing the KL divergence between prior and posterior, while the forward model trains to minimize that same term. This establishes a dynamic push-pull effect, driving self-organized exploration. 
Please note that the imitation term in Eq.~\ref{eq:G_1} depends on external demonstrations or expert policies; this term is ignored in our study, which focuses on self-exploration.

From this formulation of expected free energy, the $Q$-value can be updated as:

\begin{align}\label{eq:bellman_plus_curiosity} 
Q(t) &= r_t + \eta D_{KL}[q_\psi(w_t|o_{t+1})||q_\psi(w_t)] + \nonumber \\ 
&\gamma (1 - done_t) \mathbb{E}_{o_{t+1} \sim D, a_{t+1} \sim \pi_\phi} [Q_{\bar{\theta}}(o_{t+1}, a_{t+1})] + \alpha \mathcal{H}(\pi_{\phi} (a_{t+1} | o_{t+1}))
\end{align}

\noindent
Here, $\eta > 0$ and $\alpha > 0$ are hyperparameters weighting the intrinsic reward based on the curiosity and the motor entropy, respectively. 

In our experiments, each episode ended after 30 steps, or terminated earlier if the agent successfully executed the command. Completed episodes are stored in a recurrent replay buffer, which can hold up to 256 episodes. When the buffer is full, the buffer discards the oldest episodes to accommodate new episodes. To ensure uniform episode length, all episodes were padded to 30 steps with empty transitions. Hence, transitions are stored with the form $\{o_t, a_t, r_t, o_{t+1}, done_t, mask_t\}$, where $mask_t = 1$ for real transitions, and $mask_t = 0$ for empty transitions added for padding. After each episode, a batch of 32 episodes was sampled from the buffer and used to train the forward model, actor, and critics. During training, loss terms were multiplied by $mask_t$, removing the influence of empty transitions.

\subsection*{Details of the Model Architecture}
\label{sec:model_architecture}
\begin{figure}[ht!]
    \centering
    \includegraphics[width=.95\textwidth]{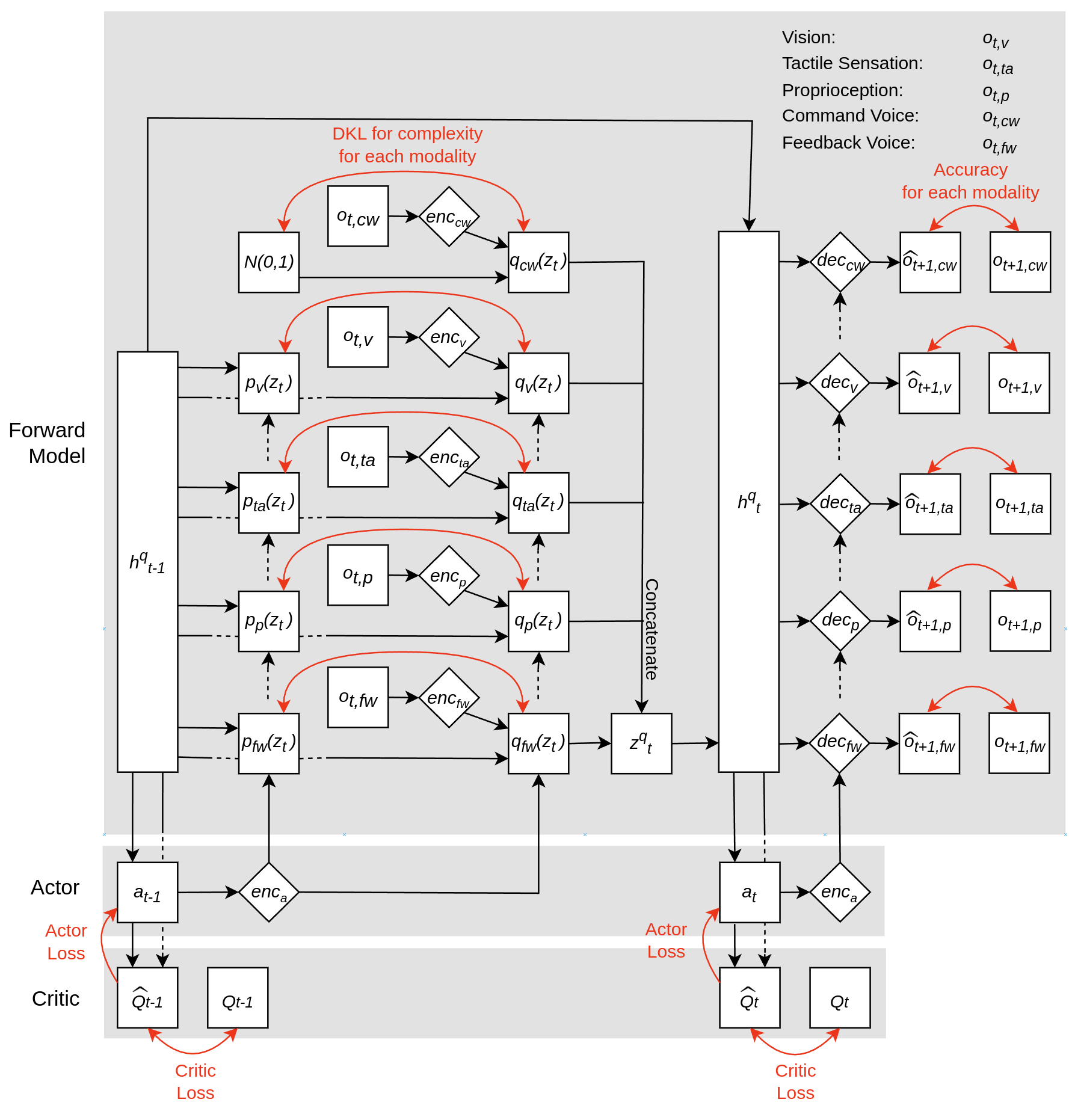}
    \caption{\textbf{The details of the proposed model architecture.}}
    \label{fig:new_forward}
\end{figure}

This subsection explains further details about the model architecture employed in this current study.
As noted earlier, the present architecture extends our previous model \cite{tinker2024intrinsic}, which is described in the ``Free energy principle, Active Inference, and Kawahara Model'' section of the Supplementary Materials.
The primary extension involves the use of separate random latent variables, encoders, and decoders for each sensory modality. This design allows the model to process multiple types of sensation independently, including vision, tactile input, proprioception, command voice, and feedback voice. 
Regarding proprioception, our model uses an encoder for the 4-dimensional motor command, which includes motor velocities for the robot's two wheels and two joint angles in its arm.
The full architecture of the proposed model is shown in Fig. \ref{fig:new_forward}.

Computation in this architecture proceeds as follows:
\begin{enumerate}
  \item The 4-dimensional motor command from the previous time step is fed into the motor command encoder, producing an encoded motor command vector.
  \item The prior distribution for each observational modality in the current time step is computed using the encoded motor command vector and the previous latent control variable. 
  Distinctly, the prior distribution for the command voice, which is constant and not impacted by the robot's motor command, is assigned a normal Gaussian distribution.
  \item The sensory observation for each modality is fed through its corresponding encoder, computing its modality-specific encoded vector.
  \item The approximated posterior distribution for each modality is approximated using its sensory encoded vector, encoded motor command vector, and the previous latent control variable. 
  Distinctly, the posterior distribution for the constant command voice is approximated based only on the encoding of the command voice alone, separating it from the motor command and latent control variable. 
  \item All posterior vectors from the current time step are concatenated across all modalities, then sampled and combined with the previous latent control variable to compute the current latent control variable.
  \item The motor command for the current time step is generated from the current latent control variable using the actor (policy network).
  \item The model predicts the next sensory observation for each modality using the current latent control variable and the current motor command, passed through the corresponding sensory decoders. Although the command voice is constant, it is still predicted; this ensures that the command voice is conserved in the latent control variable for the benefit of the actor and critic portions of the model.
  \item The $Q_t$ value is updated according to Eq. \ref{eq:bellman_plus_curiosity_method}.
  \item If the episode terminates at this step, the episode's data is saved in a recurrent replay buffer. A batch of information is sampled from the buffer to train the forward model, actor, and critic. 
\end{enumerate}
\noindent
More thorough pseudocode is provided in the following section, utilizing variables found in table \ref{table:variables}. Details of the encoders and decoders of each sensory modality (e.g., vision, tactile sensation, etcetera), as well as the motor command encoder, are described in the ``Implementation details'' section of the Supplementary Materials.

\subsection*{Pseudocode} 
The algorithm depicted in algorithm pages 1 through 3 is pseudocode representing the usage of the proposed model.

\begin{algorithm}[!htb]
\caption{Pseudocode for Proposed Model (Part One)} \label{algo:forward}
{\fontsize{12}{14}\selectfont
\begin{algorithmic}
\State Initialize forward model $f_\psi$, actor $\pi_\phi$, critic $Q_\theta$, replay buffer $R$ \Comment{Multiple critics may be used}
\State Initialize target critic weights $\bar{\theta} \leftarrow \theta$ \Comment{One target critic for each critic}
\For{epoch = 0, M}
    \Statex \Comment{In each epoch, the agent plays one episode and trains with a batch of episodes}
    \State Initialize hidden state $h^q_{-1} = 0.0$ and motor command $a_{-1} = 0.0$ \Comment{Begin new episode}
    \State Receive observation with $n$ parts, $o_0 = o_{0,0}, ..., o_{0, n}$
    \For{t = 0, T} \Comment{Steps in episode}
        \For{i = 0, n} \Comment{Parts in observation}
            \State $o^{enc}_{t,i} \leftarrow \textrm{enc}_i(o_{t,i})$ \Comment{Encode observation part}
            \State $\mu^q_{t,i}, \sigma^q_{t,i} \leftarrow \textrm{MLP}^{post}_i(h^q_{t-1} || a_{t-1} || o^{enc}_{t,i})$ 
            \Statex \Comment{Posterior inner state distribution; $||$ denotes concatenation}
            \Statex \Comment{The posterior of the constant command voice does not utilize $h^q_{t-1}$ or $a_{t-1}$}
            \State $z^q_{t,i} \sim q(z_{t,i}) = \mathcal{N}(\mu^q_{t,i},\sigma^q_{t,i}))$ \Comment{Sample posterior inner state}
        \EndFor
        \State $h^q_t \leftarrow \textrm{RNN}(h^q_{t-1}, z^q_{t,0} || ... || z^q_{t,n})$ \Comment{Advance $h^q$}
        \State Execute motor command $a_t \leftarrow \pi_\phi(h^q_t)$ to receive $o_{t+1}$, $r_t$, and $done_t$ 
        \Statex \Comment{If $done_t$, stop episode}
    \EndFor
    \State Store episode's transitions $(o_{0:T+1}, a_{-1:T}, r_{0:T}, done_{0:T})$ in $R$ \Comment{Save episode}
    \algstore{bkbreak}
\end{algorithmic}
\textit{Note:} Algorithm continues.}
\end{algorithm}

\begin{algorithm}[!htb] 
\ContinuedFloat
\caption{Pseudocode for Proposed Model (Part 2)} \label{algo:forward_2}
{\fontsize{12}{14}\selectfont
\begin{algorithmic}
    \algrestore{bkbreak}
        \State Sample batch of episodes $(o_{0:T+1}, a_{-1:T}, r_{0:T}, done_{0:T}, mask_{0:T})$ from $R$ \Comment{Sample batch}
        \State Initialize forward model hidden state $h^q_{-1} = 0.0$ \Comment{Begin training}
            \For{t = 0, T+1} \Comment{Steps in advancing forward model with batch}
                \State Initialize $P_{t-1} = 0.0$  \Comment{Begin tracking curiosity}
                \For{i = 0, n} \Comment{Parts in observation}
                    \State $\mu^p_{t,i}, \sigma^p_{t,i} \leftarrow \textrm{MLP}^{prior}_i(h^q_{t-1} || a_{t-1})$ \Comment{Prior inner state distribution}
                    \Statex \Comment{Prior of the constant command voice is assigned a normal Gaussian distribution}
                    \State $o^{enc}_{t,i} \leftarrow \textrm{enc}_i(o_{t,i})$ \Comment{Encode observation part}
                    \State $\mu^q_{t,i}, \sigma^q_{t,i} \leftarrow \textrm{MLP}^{post}_i(h^q_{t-1} || a_{t-1} || o^{enc}_{t,i})$ \Comment{Posterior inner state distribution}
                    \Statex \Comment{The posterior of the constant command voice does not utilize $h^q_{t-1}$ and $a_{t-1}$}
                    \State $P_{t-1} += \eta_i D_{KL}[q(z_{t,i})||p(z_{t,i})]$ 
                    \Statex \Comment{Compare these prior and posterior to add to curiosity}
            \State $z^q_{t,i} \sim q(z_{t,i}) = \mathcal{N}(\mu^q_{t,i},\sigma^q_{t,i}))$ \Comment{Sample posterior inner state}
        \EndFor
        \State $h^q_t \leftarrow \textrm{RNN}(h^q_{t-1}, z^q_{t,0}||...||z^q_{t,n})$ \Comment{Advance $h^q$}
    \EndFor
    \For{t = 0, T+1} \Comment{Steps in advancing critics with batch}
        \State $\widehat{Q}_t \leftarrow Q_\theta(h^q_{t-1}, a_t)$ \Comment{Predict $Q$ value}
        \State $a'_t \leftarrow \pi_\phi(h^q_t)$ \Comment{Make new motor command with actor}
        \If{$t > 1$}  \Comment{After first step, make target $Q$-values with target critics} 
            \State $\overline{Q_t} \leftarrow Q_{\bar{\theta}}(h^q_{t-1},a'_t)$ \Comment{Get target critic's $Q$ value}
            \State $Q_t \leftarrow r_t + P_t + \gamma (1-done_t) (\overline{Q_t} - \alpha\mathcal{H}(\pi_\phi(a'_t|o_t)))$ 
            \Statex \Comment{Make target $Q$ value (eq. \ref{eq:bellman_plus_curiosity})}
        \EndIf 
    \EndFor 
    \algstore{bkbreak}
\end{algorithmic}
\textit{Note:} Algorithm continues.
}
\end{algorithm}

\begin{algorithm}[!htb] 
\ContinuedFloat
\caption{Pseudocode for Proposed Model (Part 3)} \label{algo:forward_3}
{\fontsize{12}{14}\selectfont
\begin{algorithmic}
    \algrestore{bkbreak}
    \State Initialize $F = 0.0$ \Comment{Initiate free energy}
    \For{i = 0, n} \Comment{Parts in observation}
        \State $F += (\beta_i D_{KL}[q(z_{0:T,i})||p(z_{0:T,i})] - \upsilon_i \mathbb{E}_{q(z_{0:T,i})}[\log p(o_{1:T+1,i}|z_{0:T,i})])*mask_{0:T}$ 
        \Statex \Comment{Add complexity, subtract accuracy (see eq. \ref{eq:F})}
    \EndFor
    \State $\psi \leftarrow \psi - \lambda_\psi \frac{\partial F}{\partial \psi}$ \Comment{Train forward model}
    \State $J_Q(\theta) \leftarrow (\widehat{Q}_{0:T} - Q_{1:T+1})^2*mask_{0:T}$ \Comment{Critic loss}
    \State $\theta \leftarrow \theta - \lambda_Q \hat{\nabla}_\theta J_Q(\theta)$ \Comment{Train critics}
    \State $\bar{\theta} \leftarrow \tau \theta + (1 - \tau)\bar{\theta}$  \Comment{Update target critics}
    \State $J_\pi(\phi) \leftarrow (- Q_\theta (o_{0:T}, a'_{0:T}) - \alpha \mathcal{H}(\pi_{\phi}(a'_{0:T} | o_{0:T})))*mask_{0:T}$ 
    \Statex \Comment{Actor loss. Use the lowest $Q$-value among critics.}
    \State $\phi \leftarrow \phi - \lambda_\pi \hat{\nabla}_\phi J_\pi(\phi)$ \Comment{Train actor}
    \If{utilizing dynamic $\alpha$ with target entropy $\mathcal{\bar{H}}$}
        \State $J_\alpha(\alpha) \leftarrow \log (\alpha) \cdot (\mathcal{H}(\pi(a_{0:T}|o_{0:T})) - \bar{\mathcal{H}}) * mask_{0:T}$ \Comment{Alpha loss}
        \State $\log(\alpha) \leftarrow \log(\alpha) - \lambda_\alpha \hat{\nabla}_\alpha J_\alpha(\alpha)$ \Comment{Train alpha}
    \EndIf 
\EndFor
\end{algorithmic}
\textit{Note:} Algorithm is completed.
}
\end{algorithm}

\subsection*{Implementation details}

\subsubsection*{Vision}

\begin{figure}[ht!]
    \centering
    \includegraphics[width=.5\textwidth]{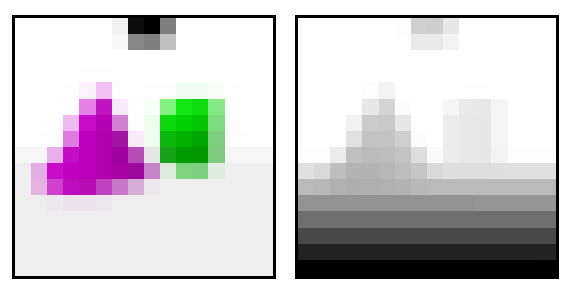}
    \caption{\textbf{The agent's vision, $\bm{o_{t,v}.}$} The robot is facing a magenta cone and a green pillar. The robot also sees part of its hand. The image on the left depicts the red, green, and blue channels. The image on the right depicts the distance.}
    \label{fig:vision}
\end{figure}

The robot visually senses the environment in the direction the robot faces with a $16\times16\times4$ image, with the four channels being red, green, blue, and distance. See Fig. \ref{fig:vision}.

In our proposed model, in order to make the approximated posterior for visual sensations, images are flattened and encoded using a linear neural network with Parametric Rectified Linear Unit activation (PReLU). 
To generate a prediction of the next image, $h^q_t$ and $a^{enc}_t$ are concatenated and decoded with another linear neural network, shaped into a $16x16x4$ tensor, and finished with a convolutional layer. See details in table \ref{table:vision}.

\begin{table}[h]
\centering
\begin{tabular}{|c|c|c|p{8cm}|}
\hline
\textbf{Layer} & \textbf{Type} & \textbf{Activation} & \textbf{Details} \\
\hline
\multicolumn{4}{|c|}{\textbf{Encoder, $enc_v$}} \\
\hline
1 & Flatten & & Shape (16, 16, 4) to shape (1024). \\
\rowcolor{lightgray}
2 & Linear & PReLU & To shape (128).\\
\hline
\multicolumn{4}{|c|}{\textbf{Decoder}, $dec_v$} \\
\hline
1 & Linear & BatchNorm2d, PReLU & From shape (264) to shape (8 * 8 * 64). \\
\rowcolor{lightgray}
2 & Reshaping & & To shape (8, 8, 64). \\
3 & CNN & Tanh & \makecell[l]{Kernel size 3, reflective padding 1. \\ To shape (8, 8, 8).}\\
\rowcolor{lightgray}
4 & Pixel Shuffle & & To shape (16, 16, 4). \\
\hline
\end{tabular}
\caption{\textbf{Encoder and decoder of agent's visual sensations, $\bm{o_{t,v}}$.}}
\label{table:vision}
\end{table}

\subsubsection*{Touch}

\begin{figure}[ht!]
    \centering
    \begin{minipage}{0.9\textwidth}
        \centering
        \includegraphics[width=\textwidth]{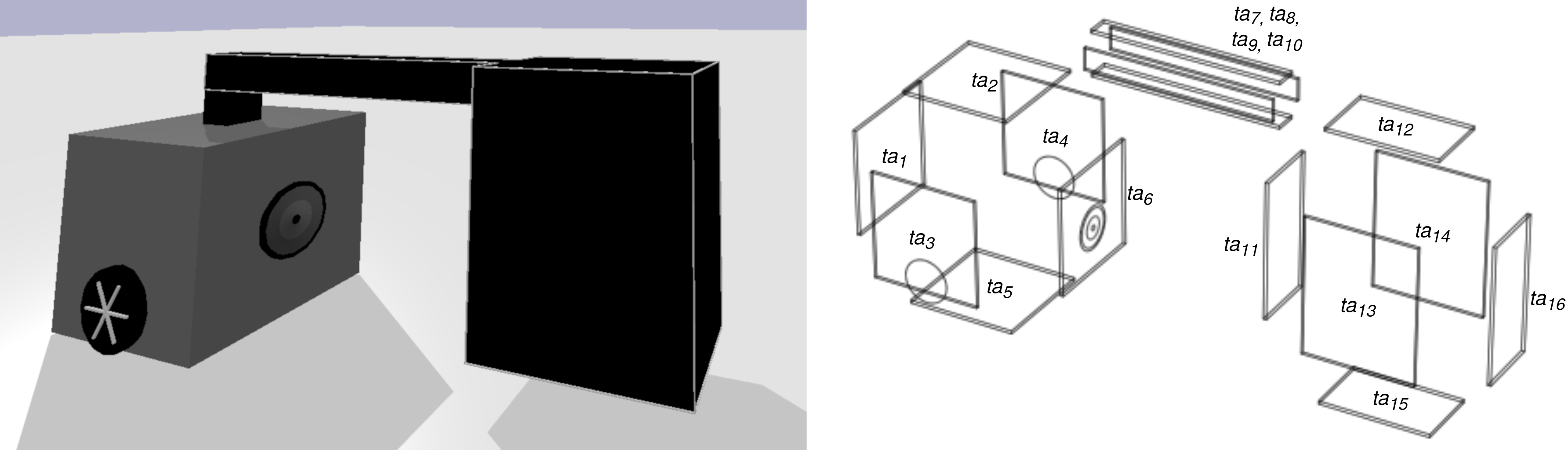}
    \end{minipage}
    \caption{\textbf{The agent's sensors for tactile sensations of touch, $\bm
    {o_{t,ta}}$.} The robot has 16 sensors, which are planes on the surface of the robot's body, arm, and hand. The camera and wheels are marked just for clarity. }
    \label{fig:sensors}
\end{figure}

The second part of the sensory observation is the tactile sensation of touch. This is represented by one value between 0 and 1 for each of the robot's 16 sensors. Each value is equal to the fraction of time in the previous step during which the respective sensor was in contact with an object. See Fig. \ref{fig:sensors}. 

In our proposed model, in order to make the approximated posterior for tactile sensation, the tensor is encoded using a linear neural network with PReLU.
To generate a prediction of the next tactile sensation, $h^q_t$ and $a^{enc}_t$ are concatenated and decoded with another linear neural network. See details in table \ref{table:touch}.

\begin{table}[h]
\centering
\begin{tabular}{|c|c|c|p{8cm}|}
\hline
\textbf{Layer} & \textbf{Type} & \textbf{Activation} & \textbf{Details} \\
\hline
\multicolumn{4}{|c|}{\textbf{Encoder,} $enc_{ta}$} \\
\hline
1 & Linear & BatchNorm2d, PReLU & From shape (16) to shape (20). \\
\hline
\multicolumn{4}{|c|}{\textbf{Decoder,} $dec_{ta}$} \\
\hline
1 & Linear & BatchNorm2d, TanH & \makecell[l]{From shape (264) to shape (16).\\Result added to 1 and divided by 2 \\for values between 0 and 1.} \\
\hline
\end{tabular}
\caption{\textbf{Encoder and decoder of agent's tactile sensations, $\bm{o_{t,ta}}$.}}
\label{table:touch}
\end{table}

\subsubsection*{Proprioception}

The third part of the sensation is the angle and velocity of the arm's joints. (The velocity of the joint may not match the robot's motor commands, because collisions with objects may restrain it.)
This consists of a tensor with four values between 0 and 1: two joint angles and two joint velocities. Each value is the normalized proportion of the respective variable between its minimum and maximum range. 

In our proposed model, in order to make the approximated posterior for sensation of proprioception, the tensor is encoded using a linear neural network with PReLU. 
To generate a prediction of the next proprioception, $h^q_t$ and $a^{enc}_t$ are concatenated and decoded with another linear neural network. See details in table \ref{table:prop}.

\begin{table}[h]
\centering
\begin{tabular}{|c|c|c|p{6cm}|}
\hline
\textbf{Layer} & \textbf{Type} & \textbf{Activation} & \textbf{Details} \\
\hline
\multicolumn{4}{|c|}{\textbf{Encoder,} $enc_{po}$} \\
\hline
1 & Linear & BatchNorm2d, PReLU & From shape (4) to shape (4). \\
\hline
\multicolumn{4}{|c|}{\textbf{Decoder,} $dec_{po}$} \\
\hline
1 & Linear & BatchNorm2d, TanH & \makecell[l]{From shape (264) to shape (4).\\Result added to 1 and divided by 2 \\for values between 0 and 1.} \\
\hline
\end{tabular}
\caption{\textbf{Encoder and decoder of agent's sensation of proprioception, $\bm{o_{t,p}}$.}}
\label{table:prop}
\end{table}

\subsubsection*{Voices}

The fourth and fifth parts of the sensation are the command voice and the tutor-feedback voice, which were described briefly in the Results section.
Both voices are sequences of one-hot vectors. Table \ref{table:words_indexes} displays the 18 words (including silence) and their indexes in the one-hot vectors. For example, the command ``Watch the Red Pillar'' is represented by

\begin{equation}
  \begin{aligned}
    [0, 1, 0, 0, 0, 0, 0, 0, 0, 0, 0, 0, 0, 0, 0, 0, 0, 0] \\
    [0, 0, 0, 0, 0, 0, 0, 1, 0, 0, 0, 0, 0, 0, 0, 0, 0, 0] \\
    [0, 0, 0, 0, 0, 0, 0, 0, 0, 0, 0, 0, 0, 1, 0, 0, 0, 0].
  \end{aligned}
\end{equation}

\noindent 
If the robot has not performed any action, then the feedback voice is only one one-hot vector indicating silence:

\begin{equation}
  \begin{aligned}
    [1, 0, 0, 0, 0, 0, 0, 0, 0, 0, 0, 0, 0, 0, 0, 0, 0, 0].
  \end{aligned}
\end{equation}

\begin{table}[]
\centering

\begin{multicols}{2} 

\begin{tabular}{|>{\raggedright}p{.8in}|>{\raggedright\arraybackslash}p{2in}|}
\hline
\multicolumn{2}{|c|}{\textbf{English Word Indexes}} \\
\hline
\multicolumn{1}{|c|}{\textbf{Index}} & 
\multicolumn{1}{|c|}{\textbf{Word}} \\
\hline
0 & (Silence) \\
\hline
1 & Watch \\
2 & Be Near \\
3 & Touch the Top \\
4 & Push Forward \\
5 & Push Left \\
6 & Push Right \\

\hline
\end{tabular}

\vspace{1em}

\begin{tabular}{|>{\raggedright}p{.8in}|>{\raggedright\arraybackslash}p{2in}|}
\hline
\multicolumn{1}{|c|}{\textbf{Index}} & 
\multicolumn{1}{|c|}{\textbf{Word}} \\
\hline
7 & Red \\
8 & Green \\
9 & Blue \\
10 & Cyan \\
11 & Magenta \\
12 & Yellow \\
\hline
13 & Pillar \\
14 & Pole \\
15 & Dumbbell \\
16 & Cone \\
17 & Hourglass \\
\hline
\end{tabular}

\end{multicols}
\caption{\textbf{English words and indexes}. The English words used and their positions in one-hot vectors. }
\label{table:words_indexes}
\end{table}

\begin{figure}[ht!]
    \centering
    \includegraphics[width=.8\textwidth]{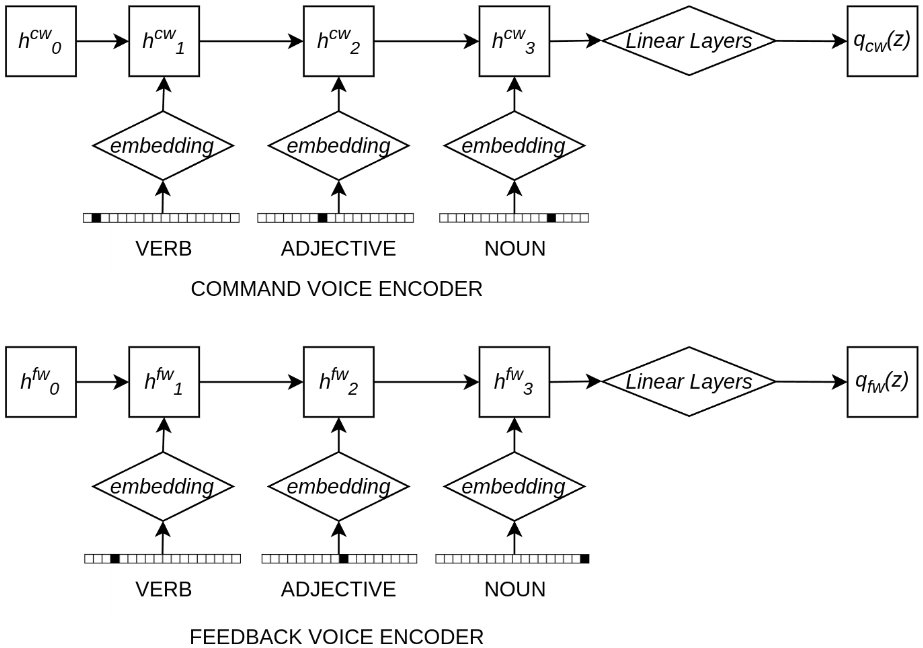}
    \caption{\textbf{Recurrent steps of command voice and feedback voice.}}
    \label{fig:voice_rnn}
\end{figure}

In our proposed model, in order to make the approximated posteriors for the sensations of each voice, the tensors are encoded using separate embeddings, recurrent neural networks, and linear layers. Note that these RNNs are ``nested'' within the forward model's RNN, such that each of the robot's steps includes three steps of interpreting each voice. 
See Fig. \ref{fig:voice_rnn}. 
To generate a prediction of the next voices, $h^q_t$ and $a^{enc}_t$ are concatenated and decoded using separate recurrent neural networks for the command voice and feedback voice. See details in table \ref{table:voice}.

\begin{table}[h]
\centering
\begin{tabular}{|c|c|c|p{8cm}|}
\hline
\textbf{Layer} & \textbf{Type} & \textbf{Activation} & \textbf{Details} \\
\hline
\multicolumn{4}{|c|}{\textbf{Encoders,} $enc_{cw}$ and $enc_{fw}$} \\
\hline
1 & Embedding & PReLU & \makecell[l]{From shape (Sequence-length, 18) \\ to shape (Sequence-length, 8).} \\
\rowcolor{lightgray}
2 & Linear & PReLU & To shape (Sequence-length, 64).\\
3 & GRU & PReLU & To shape (64). \\
\rowcolor{lightgray}
4 & Linear & PReLU & To shape (256). \\
\hline
\multicolumn{4}{|c|}{\textbf{Decoders,} $dec_{cw}$ and $dec_{fw}$} \\
\hline
1 & Linear & BatchNorm2d, PReLU & From shape (264) to shape (192). \\
\rowcolor{lightgray}
2 & Reshaping & & To shape (3, 64). \\
3 & GRU & PReLU & To shape (3, 64). \\
\rowcolor{lightgray}
4 & Linear & & To shape (3, 18). \\
\hline
\end{tabular}
\caption{\textbf{Encoder and decoder of agent's voice sensation, $\bm{o_{t,cw}}$ and $\bm{o_{t,fw}}$.}}
\label{table:voice}
\end{table}

\subsubsection*{Motor Command Encoder}

For usage in the forward model, the robot's motor commands $a_t$ are encoded into $a^{enc}_t$ with a linear neural network with PReLU. See details in table \ref{table:action}. 

\begin{table}[h]
\centering
\begin{tabular}{|c|c|c|c|c|c|c|}
\hline
\textbf{Layer} & \textbf{Type} & \textbf{Activation} & \textbf{Details} \\
\hline
\multicolumn{4}{|c|}{\textbf{Encoder, $enc_a$}} \\
\hline
1 & Linear & PReLU & From shape (4) to shape (8).\\
\hline
\end{tabular}
\caption{\textbf{Encoding motor command for forward model.}}
\label{table:action}
\end{table}

\subsubsection*{Constraints in Performing Actions}

In each step, the robot can only perform one of the six actions. This is implemented using definitions of actions and action prioritization. 
The actions ``watch,'' ``be near,'' and ``touch the top'' cannot be performed simultaneously because of requirements regarding distance from the object and touching the object. 
The actions ``push left'' and ``push right'' cannot be performed simultaneously because of the directions of movements. 
If the robot satisfies the requirements for ``touch the top,'' we reject the actions ``push forward,'' ``push left,'' or ``push right.'' 
If the robot is performing ``push forward'' and ``push left'' or ``push right,'' we accept only the action with the greatest distance pushed. 

\subsection*{Details of Experiments}

\subsubsection*{Experiment 1}
\label{sec:experiment_1_supplementary}

In Experiment 1 we trained robots with three levels of curiosity: \textit{no curiosity}, \textit{sensory-motor curiosity}, and \textit{all curiosity}. 
Table \ref{table:hyperparameters} reports the values of the hyperparameter $\eta$ for each of the four components of the sensory observations subject to exploration. These parameters represent the relative contribution of each sensory component to the robot’s intrinsic curiosity.

\begin{table}[h]
\centering
\begin{tabular}{|c|c|c|c|c|}
\hline
Name & $\eta_{vision}$ & $\eta_{touch}$ & $\eta_{proprioception}$ & $\eta_{feedback}$ \\
\hline
\hline
No Curiosity & 0 & 0 & 0 & 0\\
\hline
Sensory-Motor Curiosity & .03 & 1 & 1 & 0\\
\hline
All Curiosity & .03 & 1 & 1 & .3\\
\hline
\end{tabular}
\caption{\textbf{Hyperparameters for three types of agents.}}
\label{table:hyperparameters}
\end{table}

\begin{figure}[ht!]
    \centering
    \includegraphics[width=.4\textwidth]{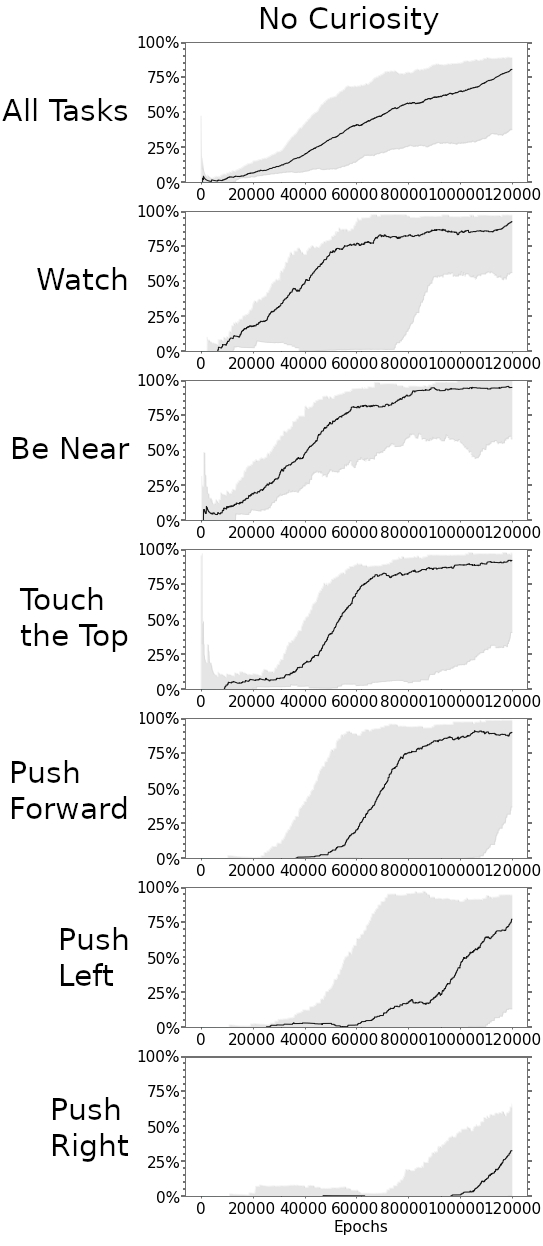}
    \caption{\textbf{Rolling success-rates over extended training in the case of \textit{no curiosity.}} Success-rates of robots with \textit{no curiosity} when trained for 120,000 epochs, which is twice the duration of robots in Fig. \ref{fig:evaluation}.}
    \label{fig:no_curiosity_120k}
\end{figure}

Fig. \ref{fig:no_curiosity_120k} shows rolling success-rates for robots in the case of \textit{no curiosity} which were trained for 120,000 epochs. Although these robots attained 80.8 percent success, it took twice the training duration of robots in the case of \textit{all curiosity} in Fig. \ref{fig:evaluation}. 

\begin{figure}[ht!]
    \centering
    \includegraphics[width=.8\textwidth]{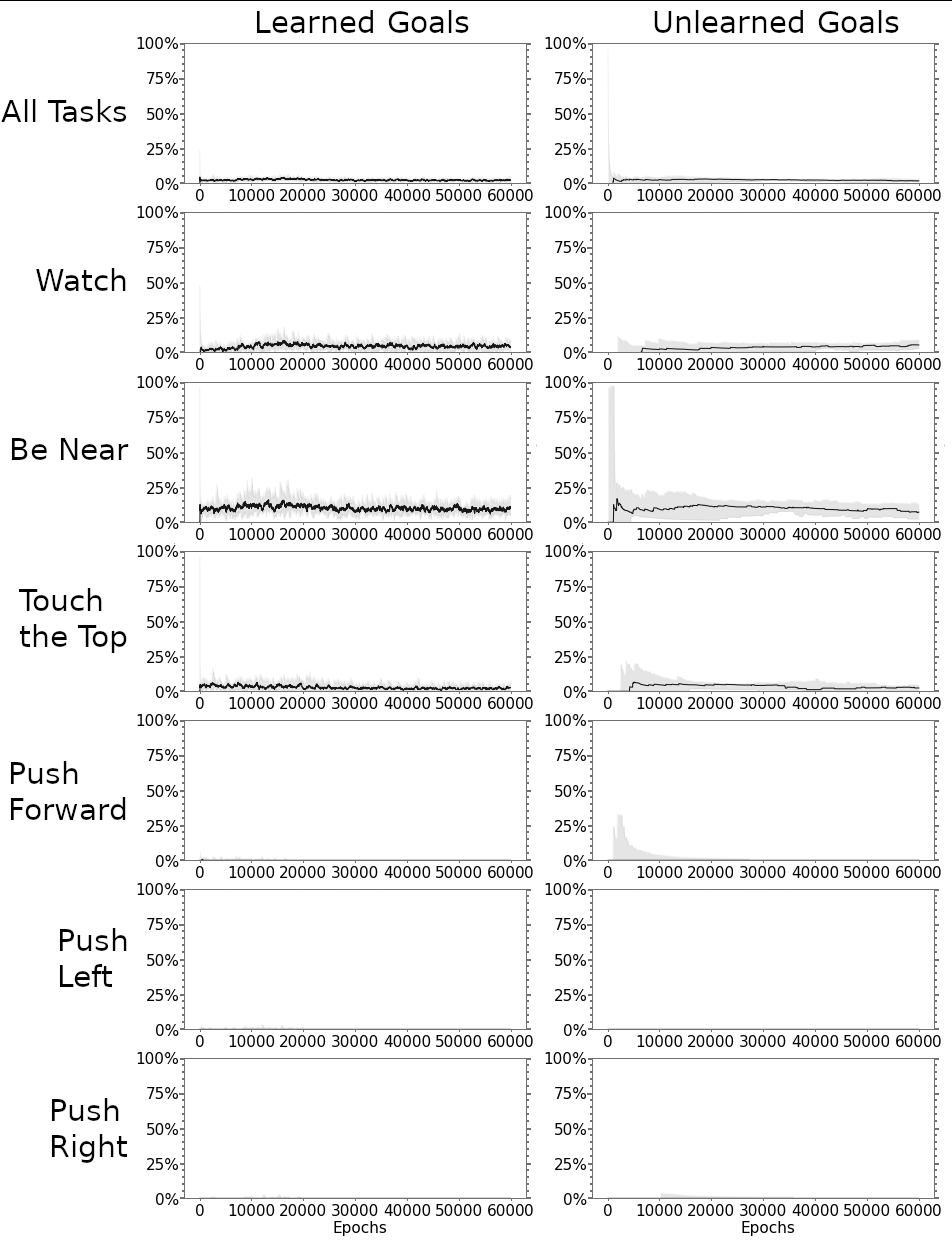}
    \caption{\textbf{Rolling success-rates for robots using the baseline recursive SAC model using all six tasks.} Poor success-rates of robots using a more traditional architecture.}
    \label{fig:baseline_results_six_tasks}
\end{figure}

\begin{figure}[ht!]
    \centering
    \includegraphics[width=.8\textwidth]{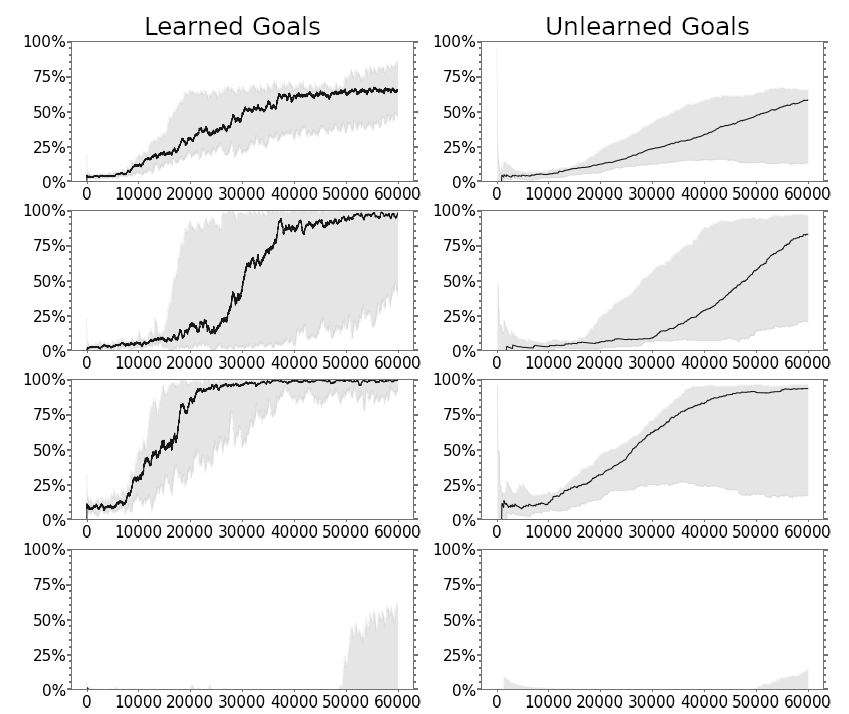}
    \caption{\textbf{Rolling success-rates for robots using the baseline recursive SAC model using just three tasks.} Success-rates of robots using a more traditional architecture.}
    \label{fig:baseline_results}
\end{figure}

Fig. \ref{fig:baseline_results_six_tasks} and Fig. \ref{fig:baseline_results} show rolling success-rates for robots utilizing a conventional baseline model architecture employing the Soft Actor-Critic (SAC) algorithm \cite{pmlr-v80-haarnoja18b} with both the actor and critics implemented with GRU \cite{recurrent}. 
This baseline model showed its best performance when the number of learning parameters was set as 2,874,013.
(Our proposed architecture has 3,998,493 learning parameters.) 
We applied the same hyperparameters for entropy used by our proposed models. When trained on all six action categories, the baseline model performed poorly, with a success-rate of nearly 0\%; see Fig. \ref{fig:baseline_results_six_tasks}. 
However, in Fig. \ref{fig:baseline_results}, the baseline model demonstrated competency with an overall success-rate of roughly 59\% with unlearned goals when trained only on the tasks ``watch,'' ``be near,'' and ``push forward.'' However, that success-rate is still worse than our proposed model utilizing curiosity-driven exploration when trained with all six tasks.

\subsection*{Statistical Analysis of U-Shaped Patterns}

To quantify U-shaped learning in exception-handling, we scored the U-shaped structure of success-rate trajectories identifying non-monotonic developmental patterns consistent with representational redescription \cite{karmiloffsmith1992beyond}. The method combines robust smoothing, normalized scaling, and piecewise isotonic regression to fit a two-phase model with a central valley.

Consider one robot's rolling-average success rate over training epochs for goals which are exceptions. The U-shape score is computed as follows:

\begin{enumerate}
    \item \textbf{Burn-in removal.} The first 10\% of training data is removed to avoid initialization noise.
    \item \textbf{Smoothing.} The curve is smoothed using a Savitzky--Golay filter with a window length of approximately 3\% of the series, reducing spurious local fluctuations.
    \item \textbf{Normalization.} The smoothed curve is linearly scaled to the $[0,1]$ range using the 5th and 95th percentiles to ensure robustness across success-rate ranges.
    \item \textbf{Valley localization.} The minimum point $i_M$ is located between 20\% and 80\% of the sequence length.
    \item \textbf{Piecewise isotonic regression.} 
    For each candidate split point $k$ near the valley (within $\pm25\%$ of the series), the left segment is fit with a decreasing isotonic regression and the right segment with an increasing isotonic regression. A cost function is minimized:
    \[
        \text{Cost}(k) = \text{MSE}(k) + \lambda \cdot (\text{drift from valley})^2 \cdot \text{MSE}_{\text{base}},
    \]
    where $\lambda = 2.0$ penalizes drifting too far from the identified valley. Indices of $k$ define the left peak $i_L$ and right peak $i_R.$
    \item \textbf{Score calculation.} 
    If the best split passes depth and width criteria (minimum 3\% depth, 6\% width), a composite U-score is computed:
    \[
        \text{U-score} = 0.6 \cdot \text{improvement} + 0.25 \cdot \text{depth} + 0.15 \cdot \text{width},
    \]
    where:
    \begin{itemize}
        \item \textit{Improvement} is the fractional MSE reduction relative to the best monotonic baseline fit.
        \item \textit{Depth} is the drop from the valley to the 90th percentile of the surrounding peaks.
        \item \textit{Width} is the relative proportion of the sequence before/after the valley.
    \end{itemize}
    \item \textbf{Index reporting.} 
    Indices of the left peak $i_L$, valley $i_M$, and right peak $i_R$ are marked with red vertical lines in Figure~\ref{fig:exceptions}.
\end{enumerate}

To compare robots trained with exceptions and without exceptions, we computed U-shape scores for each robot individually and compared the two groups using a one-tailed Welch's $t$-test (unequal variances):

\[
t = \frac{\bar{x}_1 - \bar{x}_2}{\sqrt{\frac{s_1^2}{n_1} + \frac{s_2^2}{n_2}}}.
\]

\noindent
The resulting test statistic confirmed that robots trained with exceptions exhibited significantly stronger U-shaped profiles than those without, with $p = 0.0001$.

\end{document}